\documentclass[twoside,11pt]{article}
\usepackage[preprint]{jmlr2e}
\usepackage{xcolor}
\usepackage{subcaption}
\usepackage{amsmath}

\usepackage{lastpage}

\firstpageno{1}

\begin{document}

\title{Calibrated and uncertain? Evaluating
uncertainty estimates in binary classification models}

\author{\name Aurora Grefsrud\thanks{corresponding author} \email agre@hvl.no \\
       \name Trygve Buanes \email trbu@hvl.no \\
       \addr Department of Computer science, Electrical engineering and Mathematical sciences\\
       Western Norway University of Applied Sciences\\
       5063 Bergen, Norway
       \AND
       \name Nello Blaser \email nello.blaser@uib.no \\
       \addr Department of Informatics\\
       University of Bergen\\
       5020 Bergen, Norway
       }

\editor{}

\maketitle

\begin{abstract}

Rigorous statistical methods, including parameter estimation with accompanying uncertainties, underpin the validity of scientific discovery, especially in the natural sciences. With increasingly complex data models such as deep learning techniques, uncertainty quantification has become exceedingly difficult and a plethora of techniques have been proposed.
In this case study, we use the unifying framework of approximate Bayesian inference combined with empirical tests on carefully created synthetic classification datasets to investigate qualitative properties of six different probabilistic machine learning algorithms for class probability and uncertainty estimation: (i) a neural network ensemble, (ii) neural network ensemble with conflictual loss, (iii) evidential deep learning, (iv) a single neural network with Monte Carlo Dropout, (v) Gaussian process classification and (vi) a Dirichlet process mixture model. We check if the algorithms produce uncertainty estimates which reflect commonly desired properties, such as being well calibrated and exhibiting an increase in uncertainty for out-of-distribution data points. Our results indicate that all algorithms show reasonably good calibration performance on our synthetic test sets, but none of the deep learning based algorithms provide uncertainties that consistently reflect lack of experimental evidence for out-of-distribution data points. We hope our study may serve as a clarifying example for researchers that are using or developing methods of uncertainty estimation for scientific data-driven modeling and analysis.

\end{abstract}

\section{Introduction}
Scientists deciding whether to use deep learning or other advanced statistical methods in their data analysis need to understand how reliable their results are. In classical statistical methods traditionally used in natural science research, estimated values come with uncertainties, which serve exactly this purpose. Scientists expect high uncertainties to be an indicator that they should not trust the results much, and that they may need to collect more data or re-evaluate their experiment before making any decisions or scientific claims. When we instead turn to more complex statistical methods, where the inner workings of the algorithms become unclear to scientists (making them essentially black box models), estimating uncertainty becomes both more difficult and more important.

In this study we will focus on classification problems. These are tasks where data is sorted into predefined categories based on their features, which are often assumed to be continuous variables. 
Examples in the natural sciences include determining the gender of fish based on parameters such as length and weight, categorizing patients into positive or negative for a disease based on medical test results or event classification in high energy particle physics collision experiments based on the kinematic properties of reconstructed objects. The classifier algorithms we investigate in this study estimate the probability of a new data point belonging to each of the possible classes, as well as the uncertainty of this probability.

Bayesian inference methods for classification ranging from simple parametric logistic regression to infinite dimensional nonparametric models result in probability distributions over class probabilities. Approximations of the mean of these distributions are often used as class probability estimates and the variance (or standard deviation) provide a measure of uncertainty over the space of possible solutions. Deep learning methods using stochastic gradient descent and early stopping however, typically only results in point estimates of the mode via maximum a posteriori optimization \citep{mandt2017stochastic, Duvenaud16}. Such maximization methods, similarly to maximum likelihood estimation, do not assess the variance of possible solutions, leading to lack of information about the uncertainty of the final posterior distribution. 

Various enhanced neural network classifiers aim to mitigate this issue by providing uncertainty estimates \citep{Hullermeier2021, Abdar2021}. One of the main selling points of these algorithms is that they will warn us about anomalous, also known as out-of-distribution (OOD), data because we intuitively assume that uncertainties will be high and, in the case of binary classification, probability estimates will be close to 0.5 for this kind of data. However, uncertainty estimates from ensembling, Monte Carlo Dropout and other neural network based algorithms have been shown to give arbitrarily low uncertainty estimates for OOD data \citep{Liu2021}. 

The fidelity of the in-distribution estimated class probabilities themselves has also been a topic of much debate. A well calibrated classifier outputs class probability estimates that on average match the empirical class frequency distribution. There is conflicting empirical evidence in the literature on the calibration properties of neural network classifiers. Some studies show that point estimates are not well calibrated, raising concerns about their reliability \citep{Guo2017}. Other studies show that neural networks exhibit reasonably good calibration properties \citep{Niculescu-Mizil2005, Minderer2021}. The field of uncertainty quantification in machine learning is unfortunately fraught with inconsistency, lack of theoretical justification and widespread disagreement on fundamental terms such as probability, uncertainty, and calibration, further exacerbating confusion \citep{trivedi2025, Perez-Lebel2023}.

The aim of this study is to contribute to a clearer understanding in the field of uncertainty quantification and probabilistic machine learning with a focus on binary classification. We do this by clarifying the concepts of class probability, calibration, uncertainty estimates and out-of-distribution predictions, using the theoretical framework of probability calculus and approximate Bayesian inference.  In addition, we have performed a case study where we use these theoretical insights to evaluate qualitative properties of a representative but limited subset of probabilistic learning algorithms for class probability and uncertainty estimation. Specifically, we evaluated the following algorithms: (i) a neural network ensemble \citep{mackay1995ensembles}, (ii) neural network ensemble with conflictual loss \citep{Fellaji2024}, (iii) evidential deep learning \citep{Sensoy2018}, (iv) neural network with Monte Carlo Dropout \citep{Gal2016}, (v) Gaussian process classification \citep{Rasmussen2006} and (vi) a Dirichlet Process Mixture Model \citep{Li2019}. The study investigates the estimates produced by the different algorithms using synthetic data with a clearly defined generating function with smoothly varying densities, inspired by common assumptions made in natural science experiments. Specifically, we address the following three research questions in the case of our data:
\begin{itemize}
    \item Q1: Are the estimated probabilities calibrated, that is, are they unbiased and converge towards the long-run frequency distribution?
    \item Q2: Do the uncertainty estimates go down, on average, with increasing amounts of training data?
    \item Q3: Does uncertainty increase for out-of-distribution data, that is, does it increase as we move further away from the bulk of the training data?
\end{itemize}

The research questions and relevant terminology are explained further in sections \ref{sec:theory} and \ref{sec:experiment}. The algorithms tested are described in Appendix A. We hope that our study can serve as an intuitive and easy to implement example of employing toy studies for researchers looking into developing new uncertainty quantification methods using theoretically founded methodology. We want to make it clear that the study only considers three datasets and a few of the possible uncertainty quantification methods that are available. The results of this study should not be generalized beyond this scope.

\section{Theory}\label{sec:theory}
In this section we will describe the theoretical concepts relevant to the case study. We assume the reader is familiar with basic probability theory and both Bayesian and frequentist statistics, but recap the most important aspects before explaining the concept of calibration and out-of-distribution overconfidence. Further we introduce the mean and standard deviation of the posterior distribution as our class probability and uncertainty estimates and explain how these can be calculated.

\subsection{The need for a theory of deep learning}
In the age of big-data driven modeling and decision making, where the complexity of data and statistical models continue to grow as computing power and storage capacity increases, great opportunities but also critical challenges face scientists wanting to take advantage of this new wealth of information \citep{Fan2014}. In the natural sciences, perhaps the most pressing of these issues is the need for robust and theoretically well founded statistical methods \citep{Maturo2025}. This is important because basic research in the natural sciences is especially dependent on public trust for its funding and credibility, and publishing wrong results based on faulty assumptions about the statistical methods used can potentially be devastating.

Machine learning techniques such as deep learning, a family of algorithms using artificial neural networks to
learn from training data to make predictions on unseen data, have proved to be extremely effective tools for problems involving big complex datasets. However, there has traditionally been a lack of interest in establishing mathematical theoretical foundations in the successfully empirically-driven field of deep learning \citep{breiman2001}. Despite this, as deep learning has shown itself to be exceptionally effective at solving real-life problems, significant effort has been put into post-hoc theory development as described in an overview by \cite{he2020}, who classifies attempts at theory development into six different categories. Their category ``Stochastic differential equations for modeling stochastic gradient descent and its variants" encompasses the Bayesian approach to the problem, and this is also the approach we use.

\subsection{Probability theory}

We consider the following very general model of our data. A dataset $\mathcal{D}_N :=  \{d_1, ..., d_N\}$ with \(N\) samples, where $ d_i = (\mathbf{c}_i, \mathbf{x}_i)$, is a finite sample from an infinitely exchangeable sequence\footnote{The assumption of exchangeability is slightly weaker than the more common assumption of independent and identically distributed random variables. A sequence is deemed exchangeable if we would assign it the same joint probability regardless of the order of the data points.}. We let the one-hot vector $\mathbf{c}_i$ indicate a binary\footnote{Extension to more than two classes is fairly trivial at this point, but turns out to not be as simple in practice, so we will just consider the binary case.} class label, which is considered to be a categorical random measure that can take on the values $[1,0]$ (class 1) or $[0, 1]$ (class 2). We consider the input features \(\mathbf{x}_i\) to be a collection of $d_x$ continuous real-valued random measures which can be represented by the $d_x$-dimensional vector $\mathbf{x}_i = [x_i^1, x_i^2, ... x_i^{d_x}] \in \mathbb{R}^{d_x}$. The aim of the classification problem is to calculate the predictive posterior conditional class probability distribution for a new data point, ${d_{N+1}=(\mathbf{c}, \mathbf{x})}$. This probability distribution is a binary discrete distribution given by $\mathbf{P}(\mathbf{c}|\mathbf{x}, \mathcal{D}_N) = [P(c^1 =1|\mathbf{x}, \mathcal{D}_N), P(c^2=1|\mathbf{x}, \mathcal{D}_N)]$, where ${P(c^2=1|\mathbf{x}, \mathcal{D}_N)} = 1- {P(c^1=1|\mathbf{x}, \mathcal{D}_N)}$.

By using de Finetti's representation theorem, it can be shown that if $\{\mathbf{c}_1,...,\mathbf{c}_N\}$ is a finite sample from an infinitely exchangeable sequence of measurements of the binary random quantity $\mathbf{c}_i\in \{[1,0], [0,1]\}$, then the joint probability of this data is given by the marginalization over the parameterized distribution
\begin{equation}
    \begin{split}
        P(\mathbf{c}_1,...,\mathbf{c}_N) = \int_0^1P(\mathbf{c}_1,...,\mathbf{c}_N|\nu_{\infty}=\nu)p(\nu_{\infty}=\nu)d\nu = \\
        \int_0^1 \prod_{i=1}^N{\nu}^{c^1_i}(1-\nu)^{c^2_i}p(\nu)d\nu,
    \end{split}
\end{equation}
where $0\leq \nu \leq 1$. The likelihood $P(\mathbf{c}_1,...,\mathbf{c}_N|\nu_{\infty}=\nu):=P(\mathbf{c}_1,...,\mathbf{c}_N|\nu)$ can be interpreted as the joint distribution of a set of independent measurements that are assumed sampled from a Bernoulli distribution with parameter $\nu$. The probability density distribution ${p(\nu_{\infty}=\nu):}=p(\nu)$ is the prior for the \textit{long run frequency distribution} (LRFD) of the data. For binary data the LRFD is completely specified by the scalar parameter $\nu_{\infty}=\lim_{N\rightarrow \infty}\sum_{i=1}^N c^1_i/N$. The derivations can be found in Bayesian statistics textbooks such as \cite{Bernardo}.

Assuming that the conditional class probability can be defined for every $\mathbf{x}\in\mathbb{R}^{d_x}$, we can then express the predictive posterior conditional class probability of class 1 as
\begin{equation}
    \begin{split} 
    P(c^1 =1|\mathbf{x},\mathcal{D}_N) = \int_0^1 P(c^1 =1|\nu(\mathbf{x})) p(\boldsymbol{\nu}(\mathbf{x})|\mathcal{D}_N)d\nu(\mathbf{x}) = \int_0^1 \nu(\mathbf{x}) p(\boldsymbol{\nu}(\mathbf{x})|\mathcal{D}_N)d\nu(\mathbf{x}) \\
    = \int_0^1 \nu(\mathbf{x})\frac{p(\mathcal{D}_N|\nu(\mathbf{x}))p(\nu(\mathbf{x}))}{p(\mathcal{D}_N)}d\nu(\mathbf{x})= \mathbb{E}_{\nu\sim p(\nu(\mathbf{x})|\mathcal{D}_N)}[\nu(\mathbf{x})] .
    \end{split}
    \label{eq:predictive_posterior}
\end{equation}
where $0 \leq \nu(\mathbf{x})\leq 1$. $P(c^2 =1|\mathbf{x},\mathcal{D}_N)$ is simply $1-P(c^1 =1|\mathbf{x},\mathcal{D}_N)$. 

In Bayesian statistics, the class probability is a measure of how plausible we would think it is that the data point belongs to that class. Probabilities are always conditioned on all information relevant to the problem, which may or may not include a dataset. In binary classification, $P(c^1=1|\mathbf{x})=1$ means we are completely certain that the data point belongs to class 1. An assigned probability of 0.5 means we consider it to be equally likely that it belongs to class 1 or class 2, so we are very uncertain about the classification. In frequentist/classical statistics the class probability is, by definition, the LRFD and we must estimate it empirically. The LRFD of classification problems is often referred to as the true underlying probabilities, true probabilities, real probabilities and the true posterior distribution in machine learning literature \citep{Perez-Lebel2023}. The situation gets more complicated when the class probability is taken to mean the LRFD of another process than the one actually generating the data, such as that of an ``unbiased" dataset, where the bias is unknown. We will use the epistemic Bayesian definition of probability as a measure of plausibility conditioned on the data and information we actually have, while also considering the viewpoint of the more commonly used frequentist methodology.

\subsection{Calibrated classifiers}\label{sec:calibration}
Calibration is an important frequentist concept in the uncertainty quantification literature found in machine learning research. We define a perfectly calibrated classification algorithm to be one which outputs estimated class probabilities, $P(c^1 =1|\mathbf{x},\mathcal{D}_N)$, that are equal to the LRFD, $\nu_\infty(\mathbf{x})$, as defined above. It is by definition impossible to empirically validate that a classifier has this property. The calibration literature is therefore focused on instead evaluating if a classifier's output matches the empirical class frequency distribution, $\hat{F}_N=[\hat{\nu}_N, 1-\hat{\nu}_N], \hat{\nu}_N=\sum_{i=1}^N \frac{c^1_i}{N}$, of the test set \citep{nixon2019measuring, rubin1984}.

This approach can be justified mathematically as the empirical class frequency distribution is the unbiased maximum likelihood estimator for a dataset consisting of random binary variables assumed sampled from a Bernoulli distribution. In our case, unbiased means that the expected value of the class probability estimate, with the expectation taken over all possible variations of a finite dataset, is equal to the LRFD. This can be easily derived

\begin{equation}
    \mathbb{E}_{\mathbf{c}\sim p(\mathbf{c}|\nu_\infty)} \big[ \sum_{i=1}^N \frac{c^1_i}{N}\big] =  N\sum^{c^1\in{0,1}} \frac{c^1}{N}P(c^1|\nu_\infty)=1\times\nu_\infty+0\times(\nu_\infty-1)=\nu_\infty.
\end{equation}
As we are interested in conditional class probabilities, and our test dataset is finite, we have to evaluate calibration over bins in $(P, \mathbf{x})$-space. A classifier can be judged to be locally ``well calibrated" if the sample mean of the estimated class probabilities in a specific bin are not too far from the empirical class frequency distribution calculated over the same bin.

Common tools for evaluating calibration are reliability diagrams which do the binning over $P$, and global metrics such as the expected calibration error (ECE)\citep{Guo2017} which measure the average calibration over all the bins in the reliability diagram. These may pick up on miscalibration if the data points in each bin are clustered together. Global metrics for calibration such as approximations of the Kullback-Leibler divergence and the Wasserstein distance, evaluated over the entire test set, will be dominated by in-distribution data points and are therefore not sensitive to outliers in the data.

\subsection{Asymptotic properties of calibrated classifiers}

Using calibration metrics, through $\hat{\nu}_N$, to validate the predictive posterior conditional class probabilities is equivalent to assuming that we are calculating the probabilities in Equation \ref{eq:predictive_posterior} using a specific prior called the Haldane's prior. For cases where we have a lot of data, this is an unproblematic choice which is also useful from the perspective of Bayesian statistics as we want our models to converge towards the LRFD \citep{rubin1984}. However in the cases where we only have a few data points, or none, we show that the theoretical estimates we get by using this prior does not align with the stated desired properties of uncertainty quantification.

If we do not know anything about how the classes are distributed, we should choose an uninformative prior, $p(\nu)$, that assigns equal probabilities to both classes \citep{zhu2004, jaynes03, Bernardo}. For a Bernoulli distribution the conjugate prior is the Beta distribution with density

\begin{equation*}
    \text{Beta}(\alpha, \gamma) = \frac{\Gamma(\alpha + \gamma)}{\Gamma(\alpha)\Gamma(\gamma)}\nu^{\alpha - 1}(1-\nu)^{\gamma-1}, 
\end{equation*}
where \(\Gamma\) is the Gamma function and \(\alpha, \gamma\) are its parameters. 
The resulting posterior for an experiment resulting in $N$ trials and $s_N$ observations of class 1 is also a Beta distribution with density

\begin{equation*}
    p(\nu|\mathcal{D}_N)= p(\nu|N, s_N) \propto \nu^{\alpha+s_N - 1}(1-\nu)^{\gamma+N-s_N-1}, 
\end{equation*}
and therefore 
\begin{equation*}
    P(c^1=1|\mathcal{D}_N) = \mathbb{E}_{\nu\sim p(\nu|\mathcal{D}_N)}[\nu] = \int_0^1\nu p(\nu|N, s_N) d\nu = \frac{s_N+\alpha}{N+\alpha+\gamma}.
\end{equation*}
This estimator converges to $\nu_\infty$ in the limit of infinite data, but is only unbiased if $\alpha=\gamma=0$. 

To get the desired equal class probabilities for $N=s_N=0$, we have to set $\alpha=\gamma$. 
This gives us a mean and variance of

\begin{equation}\label{eq:bernoulli_mean}
    P(c^1=1| \mathcal{D}_N)) = \mathbb{E}_{\nu\sim p(\nu|\mathcal{D}_N)}[\nu] = \frac{s_N+\alpha}{N+2\alpha},
\end{equation}
\begin{equation}\label{eq:bernoulli_var}
    \text{Var}_{\nu\sim p(\nu|\mathcal{D}_N)}[\nu] = \frac{(N-s_N+\alpha)(s_N+\alpha)}{(N+2\alpha+1)(N+2\alpha)^2}=\frac{\mathbb{E}[\nu](1-\mathbb{E}[\nu])}{(N+2\alpha+1)},
\end{equation}
where we will from now on use the shorthand notation $\mathbb{E}[\nu]$ to mean the expectation taken over the posterior distribution for $\nu$ and not over possible variations of the dataset.

As explained by \cite{zhu2004}, the choice of $\alpha$ determines how strong the influence of the prior is on the posterior.
Letting $\alpha$ go towards 0 leads to the unbiased maximum likelihood estimator $\lim_{\alpha\rightarrow0} \mathbb{E}(\nu) = \lim_{\alpha\rightarrow0}(s_N+\alpha)/(N+2\alpha)=s_N/N$. $\text{Beta}(0,0)$ is the Haldane prior and is an improper prior with all the probability mass concentrated in two point masses at 0 and 1. If we make one observation of class 1, $s_N=N=1$, and insert this data into the formula for the posterior, the resulting class probability and variance for class 1 are 1 and 0 respectively. This implies the highest possible confidence that the next datapoint will also be of class 1, contrary to our intuition for an experiment with only one datapoint.

Setting $\alpha$ to 1 specifies a flat prior over $\nu$ and gives us the Bayes-Laplace estimator $\mathbb{E}(\nu) = (1+s_N)/(2+N)$. The posterior mean of the probability of class 1 for a new data point will be \(\frac{2}{3}\), a much less extreme departure from 0.5. If we want to reduce overconfident estimates for class probability when there is little data, we should set a prior with stronger bias towards $0.5$ by choosing a larger value for $\alpha$ in the prior. If we do this, our estimates will by definition never be perfectly calibrated (unbiased in the frequentist/classical statistics sense), but the model will be less prone to high-confidence error. It is therefore necessary to separate the discussion of in-distribution (locally high training statistics) calibration, where the bias from the prior is insignificant, from the discussion of out-of-distribution (locally low training statistics) predictions, where a bias is desirable.

\subsection{Approximate Bayesian inference}
Going back to the expression for the predictive posterior in Equation \ref{eq:predictive_posterior}, we will consider $\mathbf{\nu}$ to be an analytical function $\nu(\mathbf{x})\in \mathcal{F}$, where $\mathcal{F}$ is the set containing all possible functions $\mathbf{\nu}$. We can then integrate over the in-principle infinite dimensional function-space $\mathcal{F}$ to find the mean of the posterior at any point $\mathbf{x}$
\begin{equation}\label{eq:approximate_inference}
    \mathbb{E}[\nu(\mathbf{x})] = \int_\mathcal{F} \nu(\mathbf{x}) p(\mathbf{\nu(\mathbf{x})}|\mathcal{D}_N)d\nu(\mathbf{x}) = \int_\mathcal{F} \nu(\mathbf{x})\frac{p(\mathcal{D}_N|\nu(\mathbf{x}))p(\nu(\mathbf{x}))}{p(\mathcal{D}_N)}d\nu(\mathbf{x}),
\end{equation}
where $p(\mathcal{D}_N|\nu(\mathbf{x})) = \prod_{i=1}^{N} \nu(\mathbf{x}_i)^{c^1_i}\cdot\prod_{i=1}^{N}(1-\nu(\mathbf{x}_i))^{c^2_i}$.

At this point one we can implement any problem-specific assumptions and prior knowledge to define what every possible $\nu$ means, which prior probabilities to assign to \(\nu(\mathbf{x})\), and if necessary how to approximate this integral. In scientific classification problems with real-valued input features $\mathbf{x}$, it is common to assume that $P(c^j=1|\mathbf{x})$ as a function of $\mathbf{x}$ is continuous and smooth. This is equivalent to making the assumption that the densities $p(\mathbf{x}|c^j=1)$ are continuous and smooth, as $P(c^j=1|\mathbf{x}) \propto p(\mathbf{x}|c^j=1)$.

The most common approach to inferences over function-space is to use parametric methods, which reduce the infinite-dimensional function-space by making it finite dimensional. On the other hand, nonparametric Bayesian inference methods allow us to keep the integral infinite dimensional. These latter methods are often seen as the ``gold standard" for uncertainty estimation, as they are theoretically well founded and have properties that tend to align with how we wish uncertainty to behave, such as increased uncertainty in low density regions of data space. However, implementing high dimensional nonparametric inference models is notoriously difficult and the resulting algorithms often need computationally expensive Monte Carlo simulations which make them less useful in practice. Examples of such methods are Gaussian Processes and Dirichlet Process mixture models, which make the infinite integral tractable by making assumptions which allow for using the nice analytical properties of Gaussian and Dirichlet distributions respectively. 

Traditional parametric methods assume that the data is generated by simple well-known probability distributions with just a few parameters, and inference aims to determine the value of these parameters. These methods are extremely vulnerable to model misspecification. An alternative approach to classical parametric inference is to use approximate Bayesian nonparametric inference methods such as Bayesian neural networks. In approximate Bayesian inference over function spaces, we let the functions $\nu(
\mathbf{x})$ in Equation \ref{eq:approximate_inference} be the members of a broad family of distributions $Q(\mathbf{c}|\mathbf{x}, \boldsymbol{\theta})$ represented by some high-dimensional parametrization-vector $\boldsymbol{\theta}$ with a prior distribution $p(\boldsymbol{\theta})$. We assume that this family of distributions capture most of the important variation across possible frequency distributions of the posterior \citep{Blei2016}. We can for example use a neural network with weights and biases $\boldsymbol{\theta}$ and a normalized last layer as our family of distributions $Q(\mathbf{c}|\mathbf{x}, \boldsymbol{\theta})$.
It is then possible to incorporate the information from the training dataset $\mathcal{D}_{\text{train}} = (\mathcal{C}_{\text{train}}, \mathcal{X}_{\text{train}})$ of size \(N_{\text{train}}\) to calculate the posterior probability

\begin{equation*}
    p(\boldsymbol{\theta}|\mathcal{D_\text{train}}) = 
    \frac{p(\mathcal{\mathcal{C}_{\text{train}}}|\mathcal{X}_{\text{train}}, \boldsymbol{\theta})p(\boldsymbol{\theta})}{p(\mathcal{C}_{\text{train}}|\mathcal{X}_{\text{train}})}
     = \prod^{N_\text{train}}_{i=1} \frac{Q(\mathbf{c}_i|\mathbf{x}_i, \boldsymbol{\theta})p(\boldsymbol{\theta})}
     {\int Q(\mathbf{c}_i|\mathbf{x}_i, \boldsymbol{\theta})p(\boldsymbol{\theta}) d\boldsymbol{\theta}}. 
\end{equation*}

For high dimensional $\boldsymbol{\theta}$ or analytically intractable integrals we have to estimate this integral, for example by a Monte Carlo (MC) estimate with $\{\mathbf{\theta_i}\}, i=1, 2, .., T,$ sampled from its posterior distribution $p(\boldsymbol{\theta}|\mathcal{D_\text{train}})$. This results in the approximation

\begin{equation} \label{eq:probability_MC}
    \begin{aligned}
        P(\mathbf{c}|\mathbf{x},\mathcal{D_\text{train}})  \approx \int Q(\mathbf{c}|\mathbf{x}, \boldsymbol{\theta})p(\boldsymbol{\theta}|\mathcal{D_\text{train}})d\boldsymbol{\theta} 
        =\mathbb{E}_{\theta}[Q(\mathbf{c}|\mathbf{x}, \boldsymbol{\theta})]\\
        \approx \frac{1}{T}\sum^T_{t=1} Q(\mathbf{c}|\mathbf{x}, \boldsymbol{\theta}_t) = \bar{Q}(\mathbf{c}|\mathbf{x},\mathcal{D_\text{train}} ).
    \end{aligned}
\end{equation}
The variance of the posterior distribution can also be approximated in a similar way:
\begin{equation} \label{eq:uncertainty_MC}
    \text{Var}_{\theta}[Q(\mathbf{c}|\mathbf{x}, \boldsymbol{\theta})]\approx 
    \frac{1}{T}\sum^T_t(Q(\mathbf{c}|\mathbf{x}, \boldsymbol{\theta}_t)-Q(\mathbf{c}|\mathbf{x},\mathcal{D_\text{train}}))^2 =u_Q(\mathbf{c}|\mathbf{x},\mathcal{D_\text{train}})^2
\end{equation}
This quantity measures the spread of the estimated posterior, and its square root, $u_Q$, is what we use to quantify class probability uncertainty in this study. As the inference procedure should, in the case of no approximations, make the posterior converge towards the LRFD, the uncertainty should in general decrease with larger datasets, and may approach zero. If we do not see this behavior in our model we should suspect that the model is misspecified, our approximations are too dominant, or there is some error in the implementation of the algorithm.

\section{Methodology}

In this study, we have created synthetic data representative of a generic two dimensional binary classification problem. We have specifically focused on a task with continuous, non-Gaussian, non-periodic unbounded underlying distributions of the features with significant overlap between classes.  The simplicity of the data is intentional, aiming to facilitate an in-depth examination of the algorithms rather than the data itself. Toy models are highly idealized and extremely simplified models which are a frequently used but perhaps underappreciated tool employed by natural scientists \citep{Reutlinger2018}. While toy models may not always be very useful for solving specific complex problems, they provide an excellent basis for testing basic assumptions and increase understanding. They are frequently used in statistical modeling studies but are less commonly seen in modern deep learning literature. It should be noted that the classification task is not trivial even if it is low dimensional. The class probability depends on the input features in a geometrically symmetric way, which none of the models can easily represent. The primary motivation for using toy data in this study is that by knowing the precise expression for the asymptotic joint probability distribution and the associated conditional class probabilities (referred to as the LRFD, $\nu_\infty$, in this paper) as functions of the data features, we can exactly evaluate the estimates produced by the machine learning algorithms.

We investigated six different ML classification algorithms with probability and uncertainty estimates:  neural network ensemble (NNE), neural network ensemble with conflictual loss (CL) \citep{Fellaji2024}, neural network using evidential deep learning (EDL) \citep{Sensoy2018}, neural network with Monte Carlo Dropout (MCD) \citep{Gal2016}, Gaussian Process classification (GP)\citep{Rasmussen2006} and a Dirichlet Process Mixture Model (DPMM) \citep{Li2019}. Four are deep learning algorithms while the other two are nonparametric Bayesian inference models. There exists many other algorithms for uncertainty quantification in deep learning. Some of the more well known are Bayesian neural networks with posteriors calculated using MCMC, Laplace approximations, variational inference, SWA-Gaussian and conformal predictions. It is not the goal of this paper to test all of them, but rather to create a framework for testing these kind of algorithms and evaluate a limited but representative sample of popular methods for uncertainty quantification in deep learning.
Details of the class probability and uncertainty estimation as well as extensive hyperparameter sweeps for each algorithm can be found in Appendices \ref{sec:A1} and \ref{sec:A2}. 

\subsection{Research questions and evaluation metrics}

Our first research question is if estimated probabilities are calibrated. This topic has been studied thoroughly and different metrics of calibration such as expected calibration error \citep{Naeini2015, Guo2017, Minderer2021} have been devised to evaluate calibration properties of classifiers. However, the majority of these studies are performed on real data where the LRFD is not known and must be approximated using binning schemes on finite test data as described in Section \ref{sec:calibration}. In our case, Q1 can be investigated by comparing the probability estimates for our test set (by default in-distribution data) with the LRFD of the data. Because of the symmetries of a binary classification problem where $P(c^1=1|\mathbf{x})=1-P(c^2=1|\mathbf{x})$ we will only show the results for class 1. The differences are evaluated by plotting the estimates next to the LRFD and by calculating five summary statistics of calibration on the test set, specifically

\begin{itemize}
    \item Z: The difference between the average estimated probability and the accuracy \citep{Pernot2023};
    \item WD: The Wasserstein-1 distance (WD) between the estimated distribution and the LRFD  \citep{Ramdas2017};
    \item ECE: The expected calibration error (ECE) \citep{Naeini2015}, which is a rough approximation of the WD;
    \item Mean KL-div: The mean Kullback-Leibler (KL) divergence betwen the estimated distribution and the LRFD \citep{jordan1999};
    \item LogLoss: The cross entropy loss, which is a crude Monte Carlo approximation of the non-constant term of the KL divergence \citep{BOTEV201335}.
\end{itemize}
In Appendix C we also present reliability diagrams as an extra measure of (marginal) calibration.

The second research question aims to answer if the uncertainties produced by the algorithms actually decrease when we increase the number of data points. In the study of \cite{Fellaji2024}, where ensembles with conflictual loss are introduced, the authors call this the data-related principle of epistemic uncertainty and find NNE, MCD and EDL to not follow this principle, which is the motivation for their introduction of the CL algorithm. To evaluate if this is the case for our data, we use the average of the uncertainty estimates calculated over the test set as a metric. Because of the  stochastic nature of both data and training, we expect this metric to fluctuate, but it should in general go down as we add more data to the machine learning models.

The last research question is related to generalization properties and the ability of flexible models to make sensible predictions outside the training distribution. Ideally, we would want our uncertainty estimates to be high for data which is very different from the data previously seen by the learning algorithm \citep{Hein2019}. In our case, this means that we tested to see if probability estimates approached 0.5 and uncertainty was high in the tails of the data distributions. In scientific applications it can be useful to be able to detect both that a new data point is located in the start of the tail of the distribution as well as very far away. This is especially important in safety-critical applications where both small and big shifts from what we expect to see can be problematic. We evaluate estimates for both types of OOD data by calculating the average probability estimates and uncertainty for two different extrapolation test set, one relatively near the peak of the data density and one located an order of magnitude further away. We also looked at the behavior of the same estimates in the tails of the in-distribution test set.

\section{Dataset and experiment}
\label{sec:experiment}

\begin{figure}
    \centering
    \includegraphics[width=0.9\textwidth]{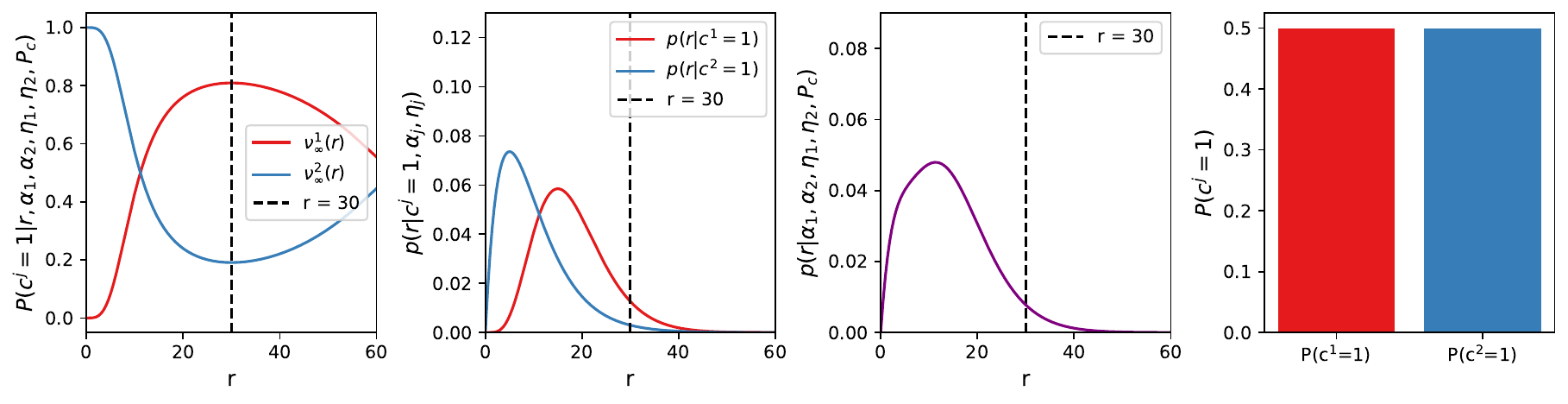}

    \includegraphics[width=0.9\textwidth]{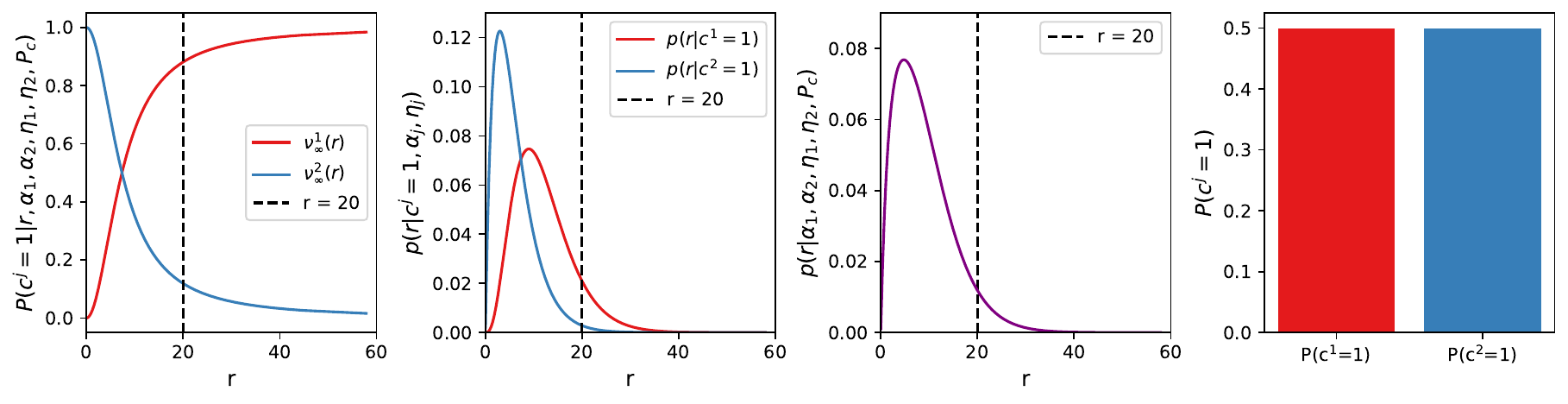}
    
    \caption{The conditional and marginal generating distributions for the two datasets. The upper row shows the distribution from which we sampled dataset A, with parameters $\alpha_{1}, \eta_{1} = [5, 2]$, $\alpha_{2}, \eta_{2} = [3, 6]$. The lower row shows the distribution from which we sampled dataset B, with parameters $\alpha_{1}, \eta_{1} = [3, 2]$ and $\alpha_{2}, \eta_{2} = [3, 4]$. The panels from left to right show the conditional distributions $P(c^j=1|r, \alpha_1, \alpha_2, \eta_1, \eta_2, P_c)=\nu^j_\infty(r)$, $p(r|c, \alpha_j, \eta_j)$, and the marginal distributions $p(r|\alpha_1, \alpha_2, \eta_1, \eta_2, P_c)$ and $P(c)$. }
    \label{fig:data}
\end{figure}

\begin{figure}
    \begin{subfigure}[b]{0.45\textwidth}
        \includegraphics[width=\textwidth]{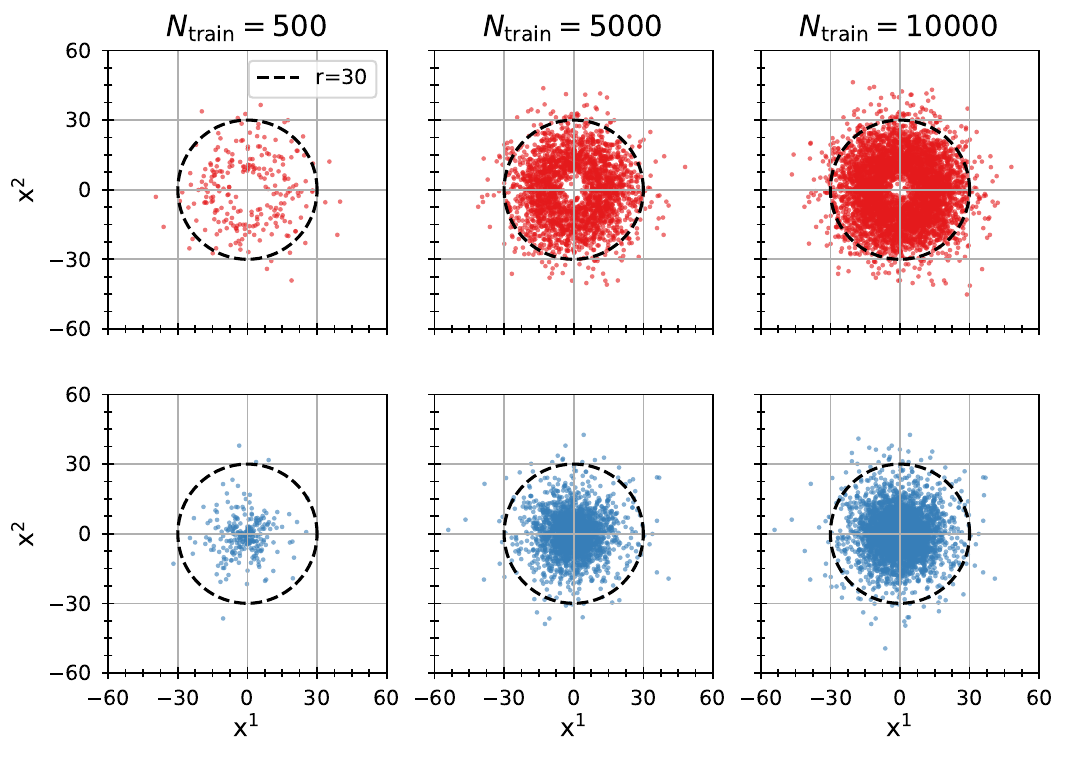}
        \caption{Training dataset A}
    \end{subfigure}
    \hfill
    \begin{subfigure}[b]{0.45\textwidth}
        \includegraphics[width=\textwidth]{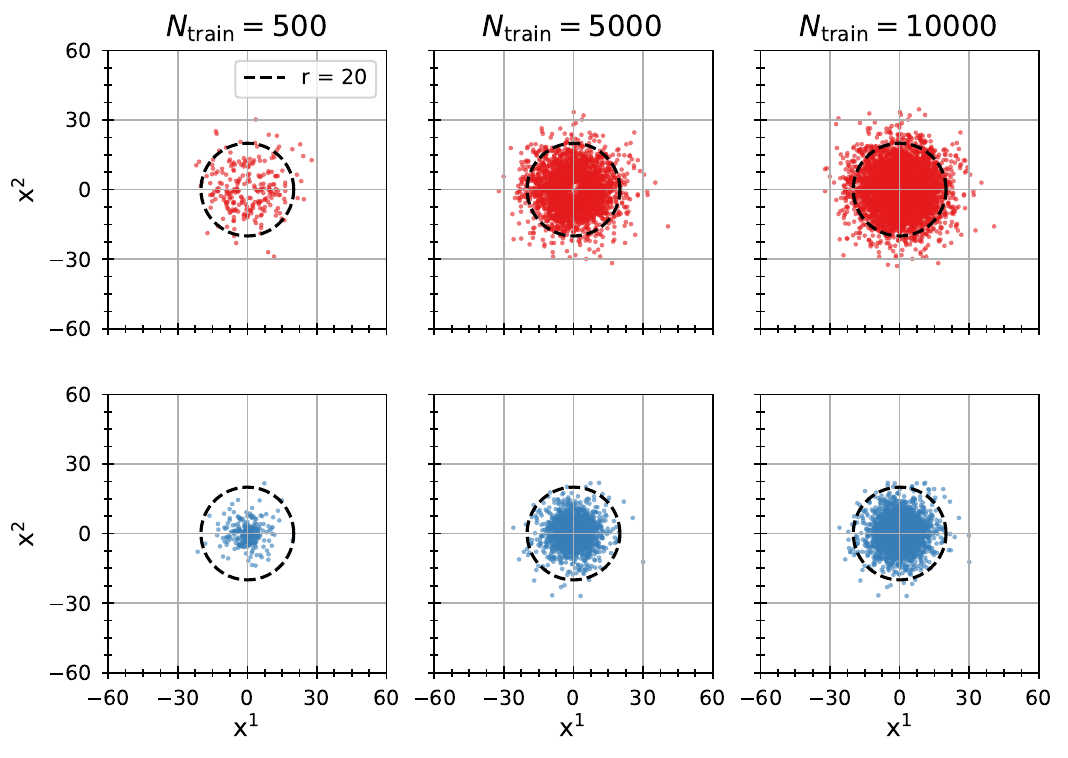}
        \caption{Training dataset B}
    \end{subfigure}
    
    \caption{Subsets of training datasets A (left) and B (right) with $N_{train}=250, 5000$ and $10000$ data points. The upper row shows the data points of class 1 in red and the lower row shows the data points of class 2 in blue.}
    \label{fig:dataset}
\end{figure}

Each data point $d$ consists of features $\mathbf{c} \in \{[1, 0],[0, 1]\}$, and $\mathbf{x} = [x^1, x^2] \in \mathbb{R}^2$ with polar coordinates $r=\sqrt{(x^1)^2+(x^2)^2}$ and $\phi = \arctan(x^2/x^1)$. To facilitate visual interpretation we have chosen a generating distribution with radial symmetry. $\mathbf{x}$ is distributed such that $p(r|c^j=1) = \Gamma(r|\alpha_j, \eta_j)$, where $\Gamma$ is the gamma distribution with shape parameter $\alpha_j$ and scale parameter $\eta_j$. The angular density is a uniform density given by $p(\phi|c^j=1)=p_\phi=1/2\pi$ for both classes. Two different classification datasets, dataset A and dataset B, were generated by setting the parameters of the gamma distributions to different values. For dataset A the parameter values are $\alpha_{1}, \eta_{1} = [5, 2]$, $\alpha_{2}, \eta_{2} = [3, 6]$ where the index indicates the class, and for dataset B the parameter values are $\alpha_1, \eta_{1} = [3, 2]$, $\alpha_{2}, \eta_{2} = [3, 4]$. The marginal class probabilities $P(\mathbf{c})=P_c=0.5$ are uniform for both datasets. The generating distributions of the data are shown in Figure \ref{fig:data}. These specific distributions were chosen because of the different behavior for $P(\mathbf{c}|r)$ in the beginning of the tails, around $r=30$ and $r=20$ for dataset A and B respectively. Train, validation and test sets with 10000, 5000, and 10000 samples, respectively, were generated by random sampling from the joint distribution $p(\mathbf{c}, x^1, x^2|\alpha_1, \alpha_2, \eta_1, \eta_2, p_\phi, P_c)$ by first sampling the class labels, then the angles and finally the radial positions of the data points\footnote{The datasets and code used for this study is made available at \hyperlink{https://github.com/choisant/foundational-modeling}{https://github.com/choisant/foundational-modeling}.}. 

To check estimated probabilities and uncertainties for OOD test points very far from the density peak of the data, a grid in polar coordinates was created by letting $\phi_i \in \{0, \frac{2}{5}\pi, \frac{4}{5}\pi, \frac{6}{5}\pi, \frac{8}{5}\pi\}$ and $r_i$ be 26 equally-spaced values in log-space for $700 \leq r \leq 1000$ for each polar angle, resulting in 130 data points. The same procedure was used to create a test set for OOD data points that were relatively close to the data, specifically in the range $ 80 \leq r \leq 120$. In the text we will refer to these two OOD test sets as "OOD-near" and "OOD-far".

Each model was trained on the same subsets of increasing amounts of training data with $N_{\text{train}} =$ 250, 500, 1000, 2000, 3000, 5000 and 10000.  Figure \ref{fig:dataset} shows the training data with 250, 1000 and 10000 training points. As the gradient descent optimization algorithm is sensitive to initialization values of its parameters and the order in which training data is seen, 20 separate training runs were performed for each NN algorithm and the one with the lowest cross entropy loss on the validation set was kept. The main results for the deep learning models are obtained using a fully connected neural network architecture and the commonly used \texttt{ReLU} activation function. The trained models were then used to estimate the class probability and uncertainty on the test set and the extrapolation grid. Details on training and implementation can be found in Appendix A. An extended study over varying hyperparameters and architecture choices are presented in Appendix B and C.

\section{Results}
\label{sec:results}

In this section we present the results of the experiment and how they relate to each research question.

\begin{figure}
    \begin{subfigure}[b]{\textwidth}
        \centering
        \includegraphics[width=0.9\textwidth]{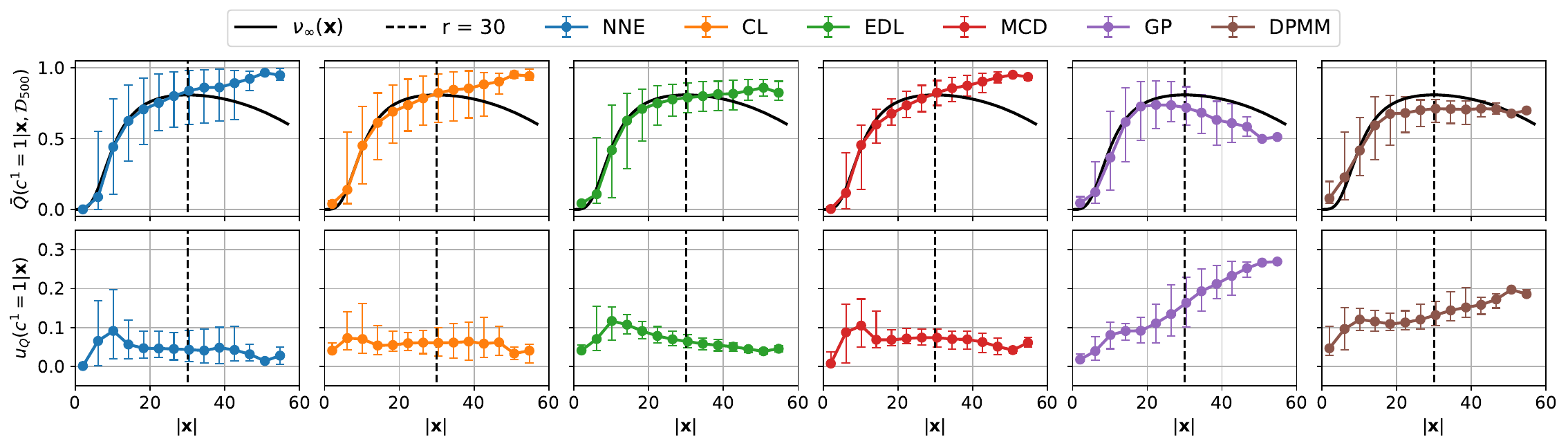}
        \caption{$N_\text{train}=500$}
    \end{subfigure}
    
    \begin{subfigure}[b]{\textwidth}
        \centering
        \includegraphics[width=0.9\textwidth]{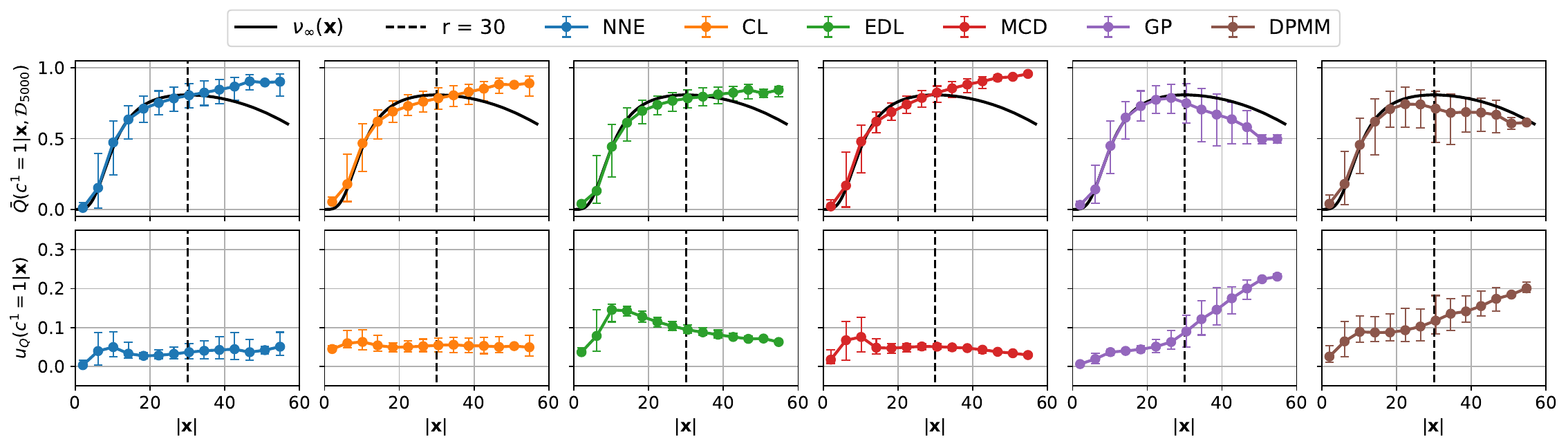}
        \caption{$N_\text{train}=5000$}
    \end{subfigure}
    
    \caption{Estimated probabilities (top row) and uncertainties (bottom row) for class 1 for the different algorithms for dataset A as a function of radius $|\mathbf{x}|$. The error bars indicate the entire spread of the data over polar angle $\phi$, while the markers indicate the sample average. The long-run frequency distribution (solid black line) is plotted for reference.}
    \label{fig:resultsA}
\end{figure}

\begin{figure}
    \begin{subfigure}[b]{\textwidth}
        \centering
        \includegraphics[width=0.9\textwidth]{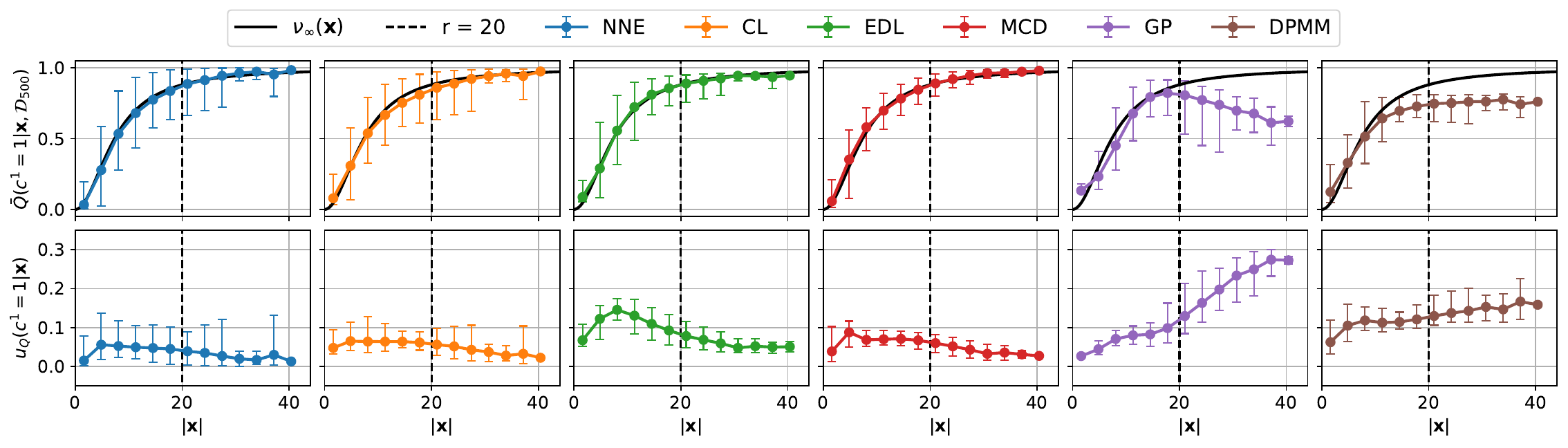}
        \caption{$N_\text{train}=500$}
    \end{subfigure}
    
    \begin{subfigure}[b]{\textwidth}
        \centering
        \includegraphics[width=0.9\textwidth]{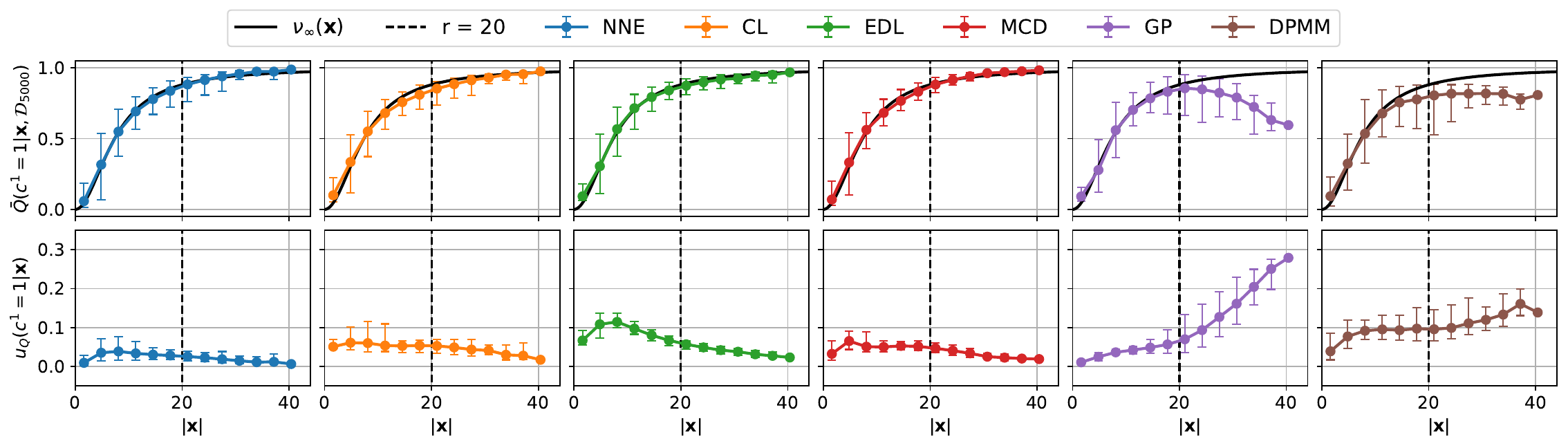}
        \caption{$N_\text{train}=5000$}
    \end{subfigure}
    
    \caption{Estimated probabilities (top row) and uncertainties (bottom row) for class 1 for the different algorithms for dataset B as a function of radius $|\mathbf{x}|$. The error bars indicate the entire spread of the data over polar angle $\phi$. The long-run frequency distribution (solid black line) is plotted for reference.}
    \label{fig:resultsB}
\end{figure}

\begin{figure}
    \centering
    \begin{subfigure}[b]{0.49\textwidth}
        \centering
        \includegraphics[width=\textwidth]{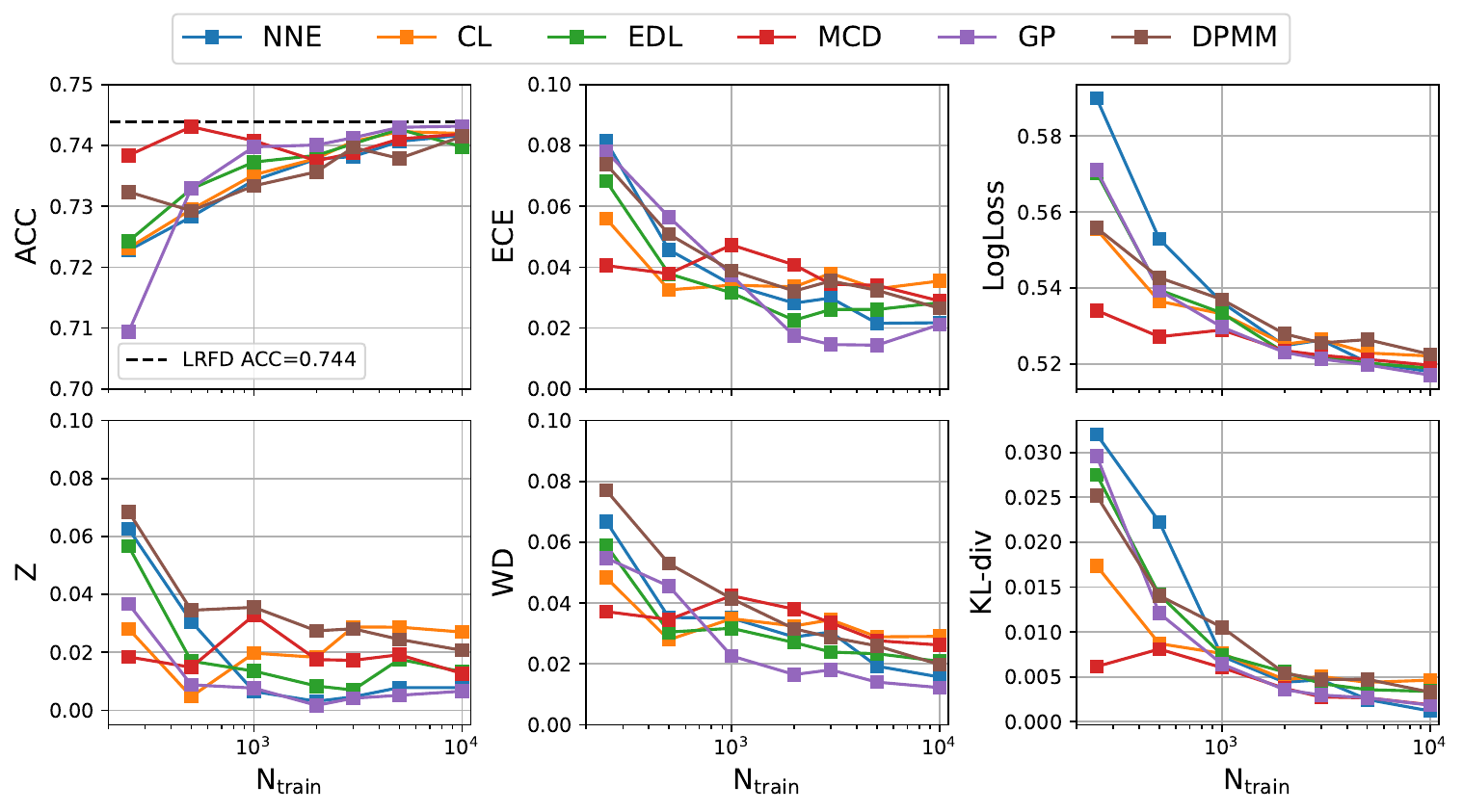}
        \caption{Dataset A}
    \end{subfigure}
    \begin{subfigure}[b]{0.49\textwidth}
        \centering
        \includegraphics[width=\textwidth]{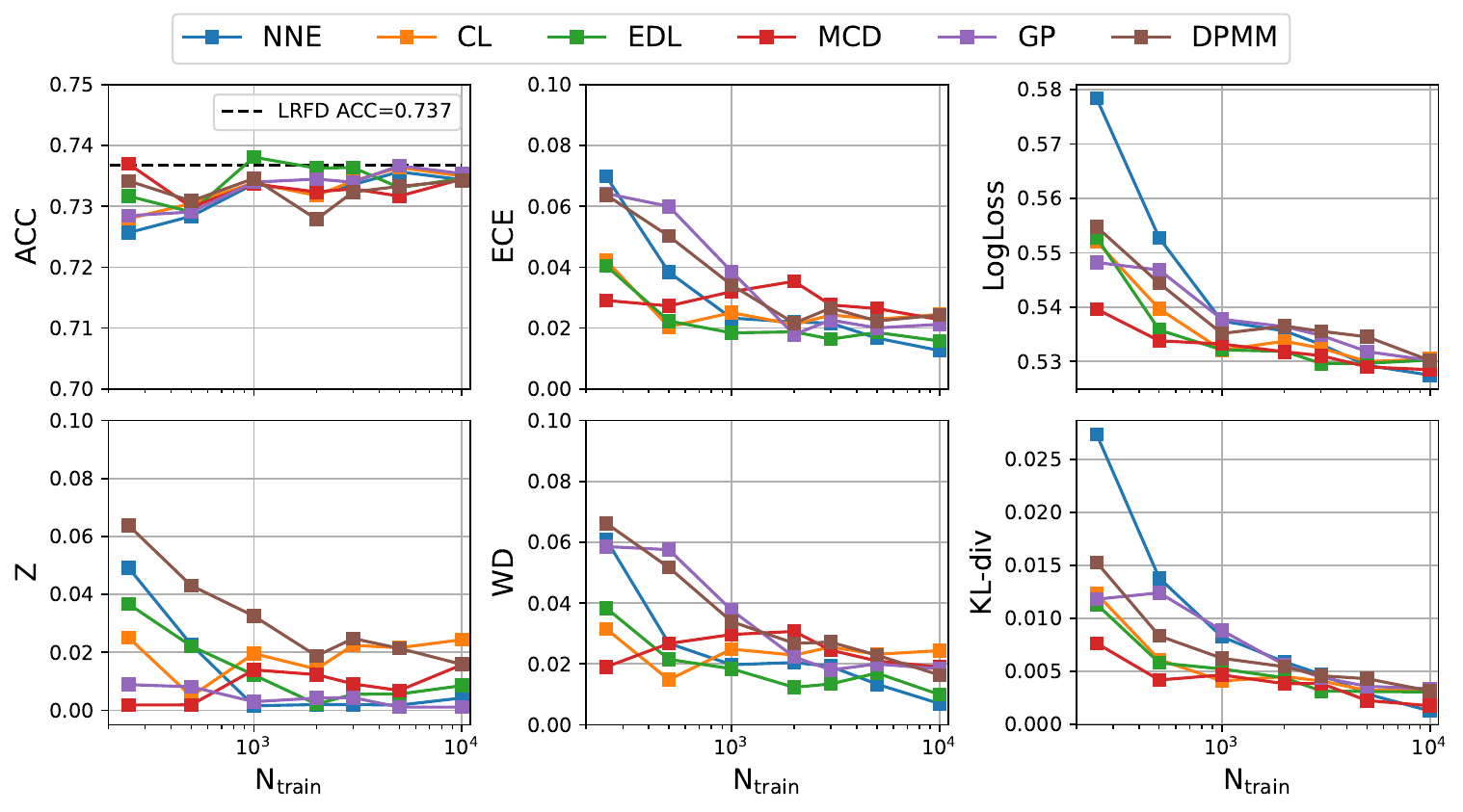}
        \caption{Dataset B}
    \end{subfigure}
    \caption{Calibration metrics of the test set as a function of number of training data points $N_\text{train}$. The six subplots show the scores of the different models: neural network ensemble (NNE, blue line), neural network ensemble with conflictual loss (CL, orange line), neural network using evidential deep learning (EDL, green line), neural network with Monte Carlo Dropout (MCD, red line), Gaussian Process classification (GP, purple line) and a Dirichlet Process Mixture Model (DPMM, brown line). The metrics calculated are the accuracy (ACC), estimated calibration error (ECE), cross entropy loss (LogLoss), model calibration error (Z), Wasserstein-1 distance (WD) and Kullback-Leibler divergence (KL-div). The dashed black line in the top left plot indicates the optimal accuracy of the test set.}
    \label{fig:metrics}
\end{figure}

\begin{figure}
    \centering
    \begin{subfigure}[b]{0.49\textwidth}
        \centering
        \includegraphics[width=\textwidth]{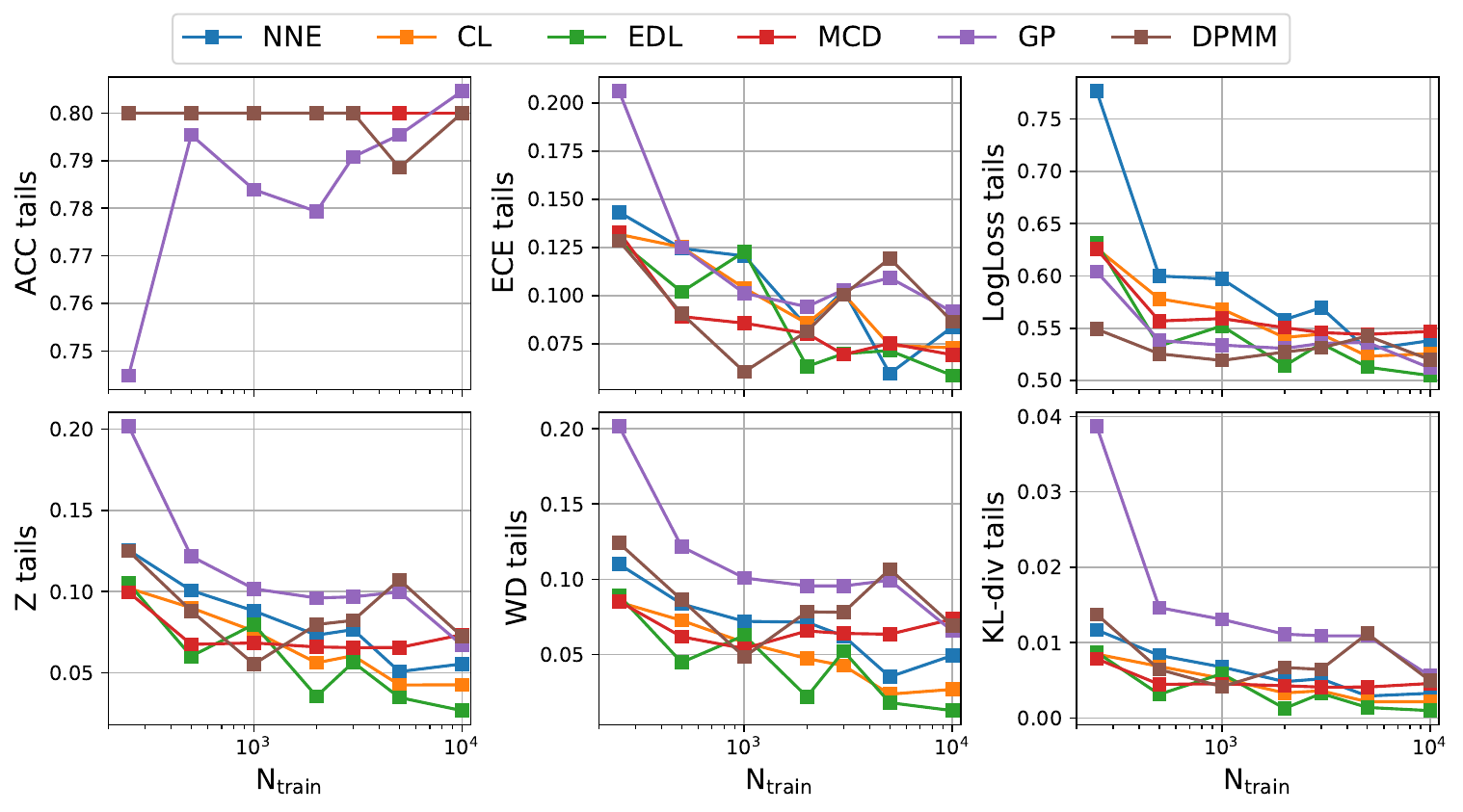}
        \caption{Dataset A}
    \end{subfigure}
    \begin{subfigure}[b]{0.49\textwidth}
        \centering
        \includegraphics[width=\textwidth]{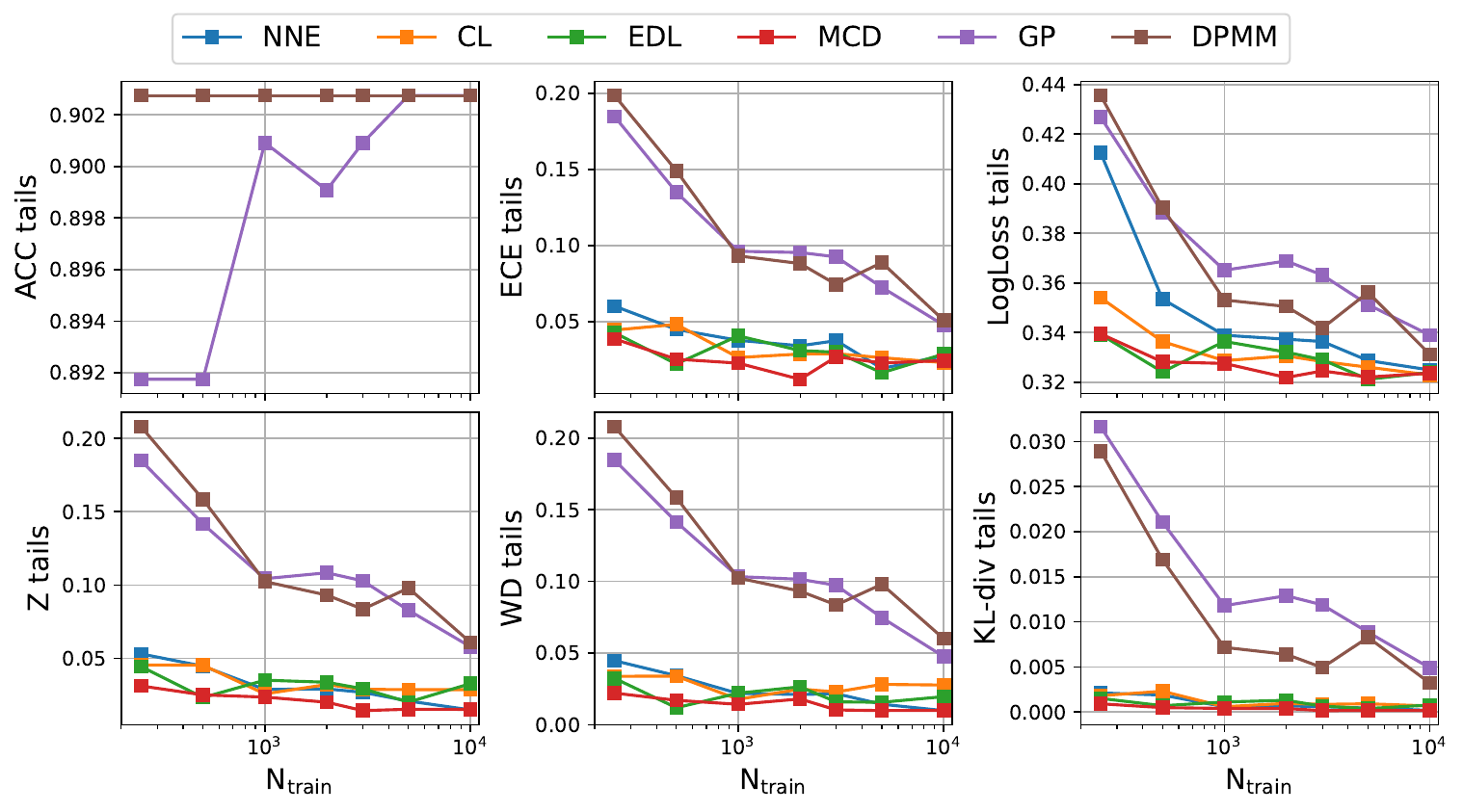}
        \caption{Dataset B}
    \end{subfigure}
    \caption{
    Calibration metrics of the test set as a function of number of training data points $N_\text{train}$ (as in Figure \ref{fig:metrics}), but calculated for the tails ($|\mathbf{x}|>30$ for dataset A and $|\mathbf{x}|>20$ for dataset B) of the test sets, as indicated by the dashed black line in Figure~\ref{fig:dataset}.}
    \label{fig:metrics_tails}
\end{figure}

\subsection{Q1: Calibration}
First we attempt to answer Q1 by evaluating if the estimated probabilities are calibrated. We find that the probability estimates for all the different models are reasonably well calibrated with respect to input features for the majority of test data points, with calibration getting better for larger values of $N_\text{train}$. In the tails of the test data distribution, the algorithms are not well calibrated unless the LRFD aligns with the extrapolation behavior of the algorithm. Global calibration metrics and reliability diagrams support this statement quantitatively.

Figure \ref{fig:resultsA} and Figure \ref{fig:resultsB} shows the estimated probabilities $\bar{Q}(c^1=1|\mathbf{x})$ and uncertainties $u_Q(c^1=1|\mathbf{x})$ as a function of polar coordinate radius $r=|\mathbf{x}|$ with error bars indicating the entire spread of estimates over polar angle $\phi$, calculated on the test set of each of the six algorithms from training on 500 and 5000 training data points from dataset A and B respectively. The figures show that probability estimates are reasonably well calibrated for all algorithms from around $0 \leq |\mathbf{x}| \leq 30$ for dataset A and $0 \leq |\mathbf{x}| \leq 20$ for dataset B. As we add more training data, the estimates concentrate around the LRFD and the error bars shrink. From around $|\mathbf{x}| \geq 30$ and $|\mathbf{x}| \geq 20$ for dataset A and B respectively, some of the estimates do not fluctuate around the LRFD and are no longer calibrated, which becomes more apparent as we move further away from the bulk of the training data. The behavior in the tails is further covered in the results of Q3.

Figure \ref{fig:metrics} and \ref{fig:metrics_tails} shows the five different calibration-related summary statistics as well as the global accuracy calculated on the test set and the data points in the tails of the test set as a function of increasing number of training data points. The plots show that all models converge towards the LRFD and the empirical estimates of the LRFD as the data size increases, but the metrics are quite noisy and do not always give the same relative ranking to each model. The neural networks are trained specifically to minimize the cross entropy loss and a decrease in this metric does not always correspond to a decrease in the other metrics. For dataset A, the GP model scores for the entire test set are better relative to many of the other models. For dataset B the opposite is true. Panel b) in Figure \ref{fig:metrics_tails} shows how calibration scores get worse in the tails for the conservative estimates of the GP and DPMM models. However, this is not necessarily reflected in the scores averaged over the entire dataset as there are naturally much fewer data points in the tails. Reliability diagrams are presented in Appendix D and show that the algorithms, when given enough data, are all well calibrated for bins over $\bar{Q}(c^1=1|\mathbf{x}) \in [0,1]$ which do not just include data points in the tails.

\begin{figure}
    \centering
    \begin{subfigure}[b]{\textwidth}
        \centering
        \includegraphics[width=0.8\textwidth]{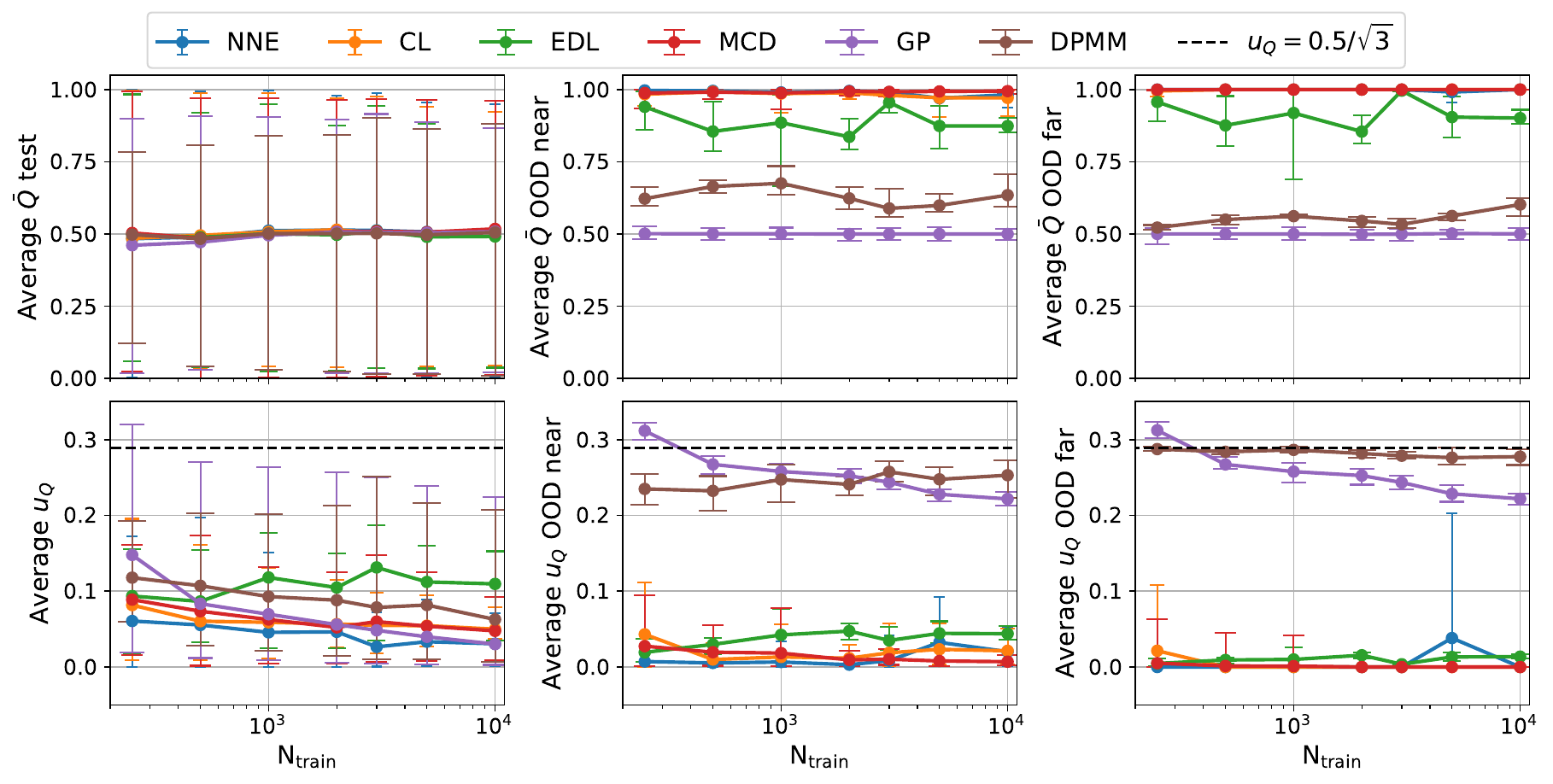}
        \caption{Dataset A}
    \end{subfigure}
    \begin{subfigure}[b]{\textwidth}
        \centering
        \includegraphics[width=0.8\textwidth]{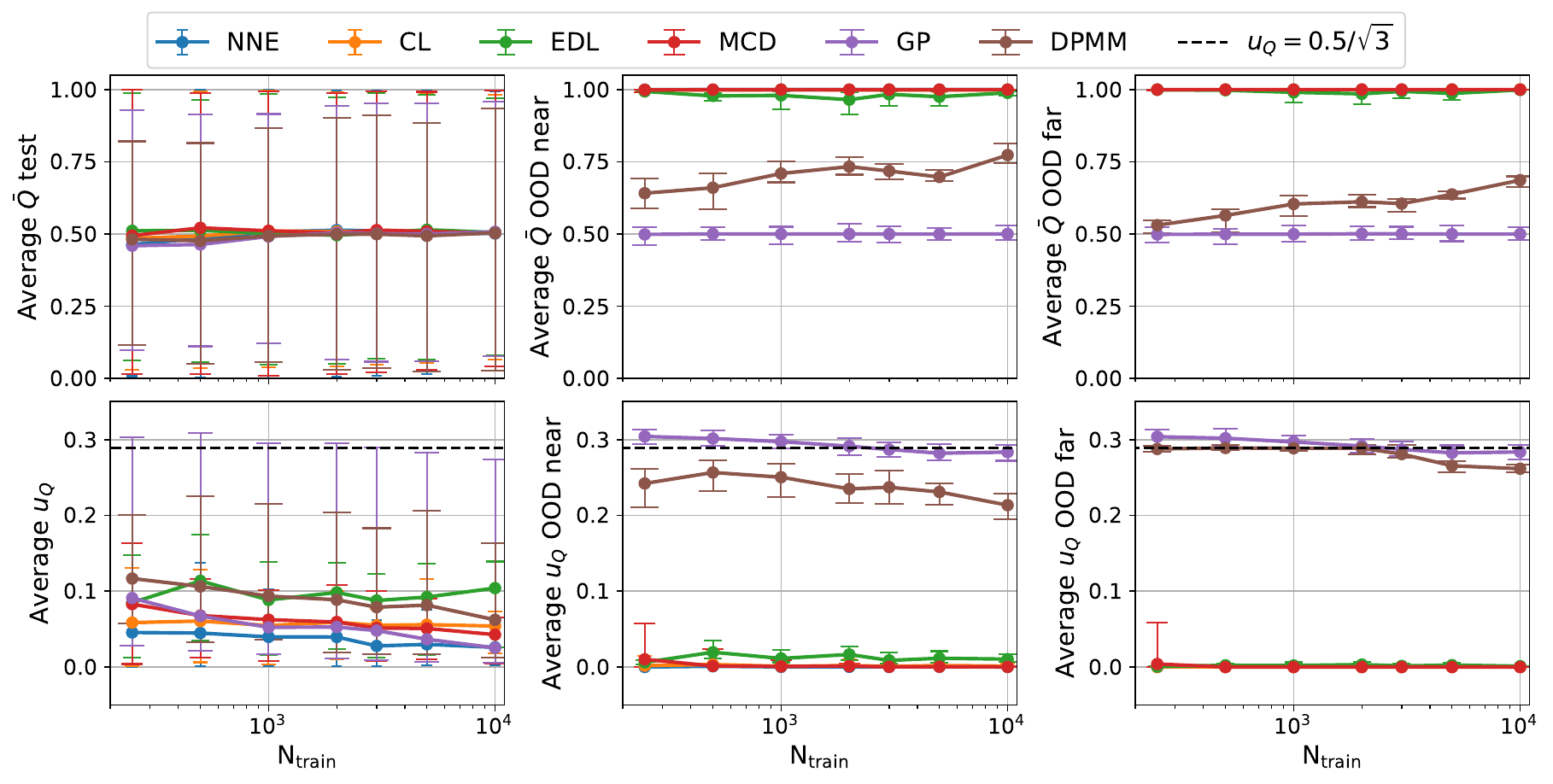}
        \caption{Dataset B}
    \end{subfigure}
    
    \caption{Top: Average estimated out-of-distribution probabilities (top rows) and uncertainties (bottom rows) as a function of $N_\text{train}$ for the test set (left), the OOD-near test set (middle) and the OOD-far test set (right). The error bars indicate the entire spread of the data over polar angle $\phi$. The colors are similar to Figure \ref{fig:metrics}. The dashed black line in the plots in the bottom rows indicates the standard deviation of the predictive posterior of a binomial model with $N=0$ and prior $\text{Beta}(1,1)$.}
    \label{fig:uncertaintiesAB}
\end{figure}

\subsection{Q2: Uncertainty vs number of training points}
Next we investigate Q2 by checking if the class probability uncertainty estimates go down as we increase the training data size. The results indicate that the average uncertainty evaluated over the test set is either constant or goes down for all models except EDL which exhibits an increase in uncertainty. For the OOD data we see the same trends, but on the OOD-near test set for dataset A we also see an increase in uncertainty for DPMM.

Figure \ref{fig:uncertaintiesAB} shows the average estimated probabilities and uncertainties for the test set, the OOD-near test set ($ 80 \leq r \leq 120$) and the OOD-far test set ($ 700 \leq r \leq 1000$). For all algorithms except EDL, the uncertainties on the test set tend to go down or remain constant as we increase the size of the training set. For EDL however, the uncertainties increase. For the OOD test points, all deep learning algorithms except EDL produce approximately constant uncertainties with value 0. The EDL uncertainties increase as a function of $N_{\text{train}}$ for the OOD-near test set for dataset A, and are approximately zero and constant for the rest. GP uncertainties decrease, which is due to the calculated approximate optimal value for the output variance of the Radial Basis Kernel decreasing with increasing nr of data points (see Appendix A). DPMM uncertainties decrease as a function of $N_{\text{train}}$ for the test sets and for the OOD test sets for dataset B, but increase for the OOD test sets for dataset A. The DPMM uncertainties may exhibit this increase due to numerical errors because of increasingly small probability density estimates in the tail, as expanded on in the next results section.

\subsection{Q3: OOD uncertainty}

Finally we answer Q3 by checking if uncertainty estimates increase as we move away from the bulk of the training data. Our main findings are that for the deep learning models using the \texttt{ReLU} activation function, uncertainties are not higher in the tails of the test set or in the OOD-near and OOD-far datasets. For the nonparametric Bayesian models, uncertainties were higher for OOD data, but the DPMM shows some non-intuitive behavior. In an additional experiment using completely identical data distributions, only the DPMM produced probability and uncertainty estimates that had high uncertainty for OOD data.

We go back to examining Figure \ref{fig:resultsA} and Figure \ref{fig:resultsB}, with a focus on the tails, defined loosely to be $|\mathbf{x}| \geq 30$ and $|\mathbf{x}| \geq 20$ for dataset A and B respectively. For dataset A, the neural network-based algorithms tend to first underestimate the class 1 probabilities, then overestimate them. For dataset B, they almost perfectly align with the LRFD. The nonparametric Bayesian inference algorithms on the other hand underestimate the probabilities in the tails relative to the LRFD for both datasets, giving class probability estimates closer to the noninformative prior probability estimate of 0.5.  In the tails of the test set distribution, class probability uncertainties increase for the nonparametric Bayesian models. The neural network models have flat or decreasing uncertainties in the tail. The uncertainty estimates of all algorithms approximately follow the relation $\sigma_\nu \propto\sqrt{\bar{\nu}(1-\bar{\nu})}$, which results in a local maxima around $\nu=0.5$. The height of this maxima is different for all the models, indicating that the uncertainty values are highly model dependent.

The first two panels of Figure \ref{fig:uncertaintiesAB} show the average estimated probabilities and uncertainties evaluated over the OOD datasets. For most deep learning models, the probability estimates are approximately 1. The only exception is EDL which for dataset A has very noisy estimates centered around 0.9. The NN-algorithms all also have estimated uncertainties of approximately 0. This is the correct value for a standard deviation proportional to $\sqrt{\nu(1-\nu)}$ when $\nu$ is 0 or 1, and so does not indicate that there is some fundamental error in the approximation of the variance of the distribution. The exception is the NNE uncertainty estimates for dataset A $N_\text{train}=5000$ which are non zero and extremely noisy. This anomaly corresponds to diverging estimates in the ensemble, which seems to be a rather unusual result when the ensemble is trained well. The NN models therefore do not show a strong, consistent increase in uncertainty with decreasing density $p(\mathbf{x})$. The GP algorithm produces probability estimates of approximately 0.5 and high uncertainty for OOD data points for both datasets independent of data size. The DPMM algorithm also produces high uncertainty for OOD data and OOD probability estimates close to 0.5 for the smallest amount of training data, but the probability estimates diverge from 0.5 as we increase the number of data points, pushing the estimated distribution towards 1. This is not what we would expect to see as the DPMM uses a Bernoulli distribution with a Bayes-Laplace (uniform) prior, which should give posterior class probabilities of 0.5 where there is no data. To calculate the conditional distribution however, the algorithm actually approximates the joint distribution, which gives very small probabilities for the data in low density areas. The results may therefore be influenced by biased numerical errors or possibly pathologies of the underlying mathematical framework \citep{Dunson2011}. The inspected MCMC diagnostics did not indicate any problems, and the behavior persisted if we changed the hyperparameter of the Beta-prior, indicating that this is a stable numerical issue of the implementation.

To study the effect of hyperparameter choice on our deep learning results we have performed a grid search over relevant hyperparameters such as the size of the neural networks, the learning rate, the weight decay and some algorithm specific hyperparameters. Detailed results are presented in Appendix B. We find that in \texttt{ReLU} neural networks some settings of hyperparameters give rise to non-extreme OOD probability estimates, but these correspond to networks that have not learned well and perform badly on the in-distribution test set. The majority of the trained networks result in extreme probability estimates similar to the ones presented in the main part of the study. Some of the hyperparameter choices also give rise to higher OOD uncertainty than in the main findings. These are also connected to networks that perform worse and often significantly so. We also tested changing the activation function in the neural networks to \texttt{tanh}, with the results presented in Appendix C. In this case, the NNE, CL and EDL algorithms sometimes produced uncertainties that increased for OOD data without sacrificing accuracy. However, the behavior was not consistent across the datasets or number of training data points. All in all, our grid search corroborates our main findings and provide some interesting additional information on the behavior of the algorithms.


\begin{figure}
    \begin{subfigure}[b]{0.45\textwidth}
        \includegraphics[width=\textwidth]{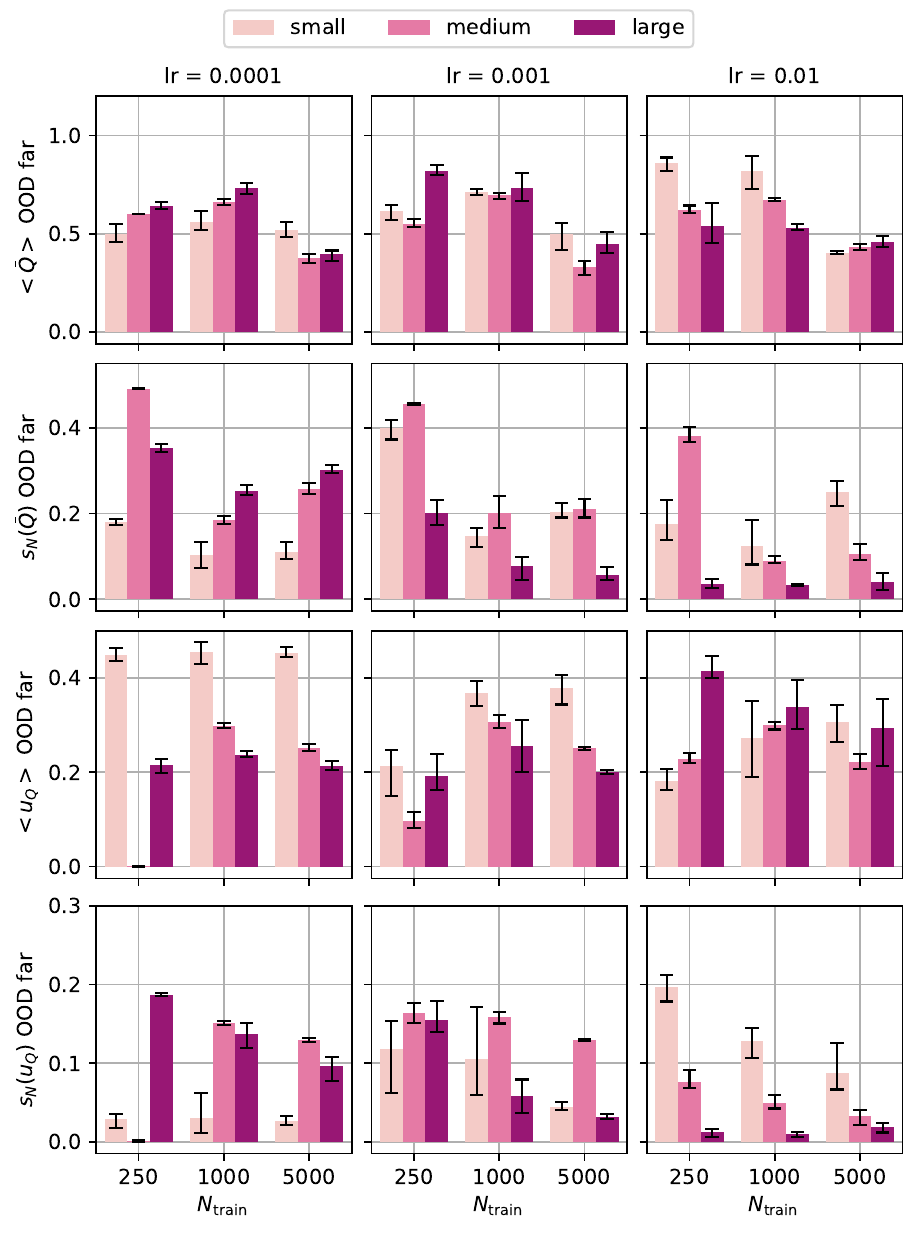}
        \caption{Neural network ensemble}
    \end{subfigure}
    \hfill
    \begin{subfigure}[b]{0.45\textwidth}
        \includegraphics[width=\textwidth]{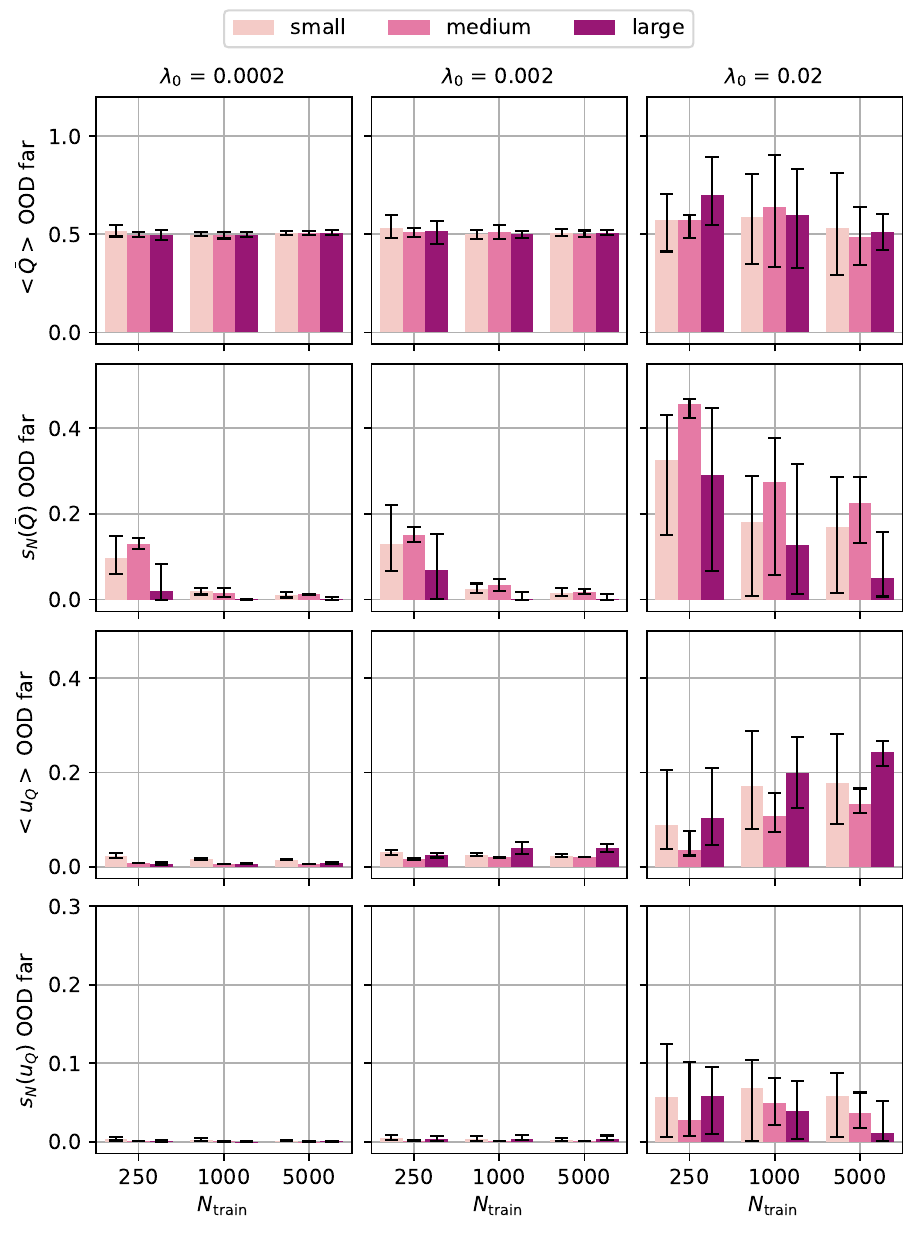}
        \caption{Evidential deep learning}
    \end{subfigure}
    
    \caption{Results for different hyperparameter settings for neural network ensemble and evidential deep learning. The columns contain the results for learning rates 0.0001, 0.001 and 0.01. The color of the bars indicates the size and depth of the hidden layers of the network. Small, medium and large has 1, 3, and 8 hidden layers with 20, 200 and 2000 nodes respectively. The first and second rows of plots shows the sample mean and sample standard deviation} of the estimated class probability for the out-of-distribution test points. The bottom two rows of plots shows the sample mean and sample standard deviation of the estimated uncertainty for the out-of-distribution test points. The error bars indicate the 2.5-97.5 percentiles.
    \label{fig:gridsearch_datasetC}
\end{figure}

As an additional test for OOD performance we created a dataset with two statistically identical classes by using the parameters of class 1 in dataset A when generating data for both classes. We ran the same hyperparameter grid search for all deep learning models. The results for NNE and EDL are presented in Figure \ref{fig:gridsearch_datasetC}. NNE, CL and MCD all produced OOD class probability estimates which varied greatly both across hyperparameter values and as a function of polar angle. EDL stood out as the only algorithm, which (for small enough $\lambda_0$) produced estimated probabilities of 0.5. Interestingly, the uncertainty was still close to zero, which means it can not be used to detect OOD data and could lead to overconfident estimates. The GP and DPMM algorithms struggled with this dataset, especially for large numbers of data points so we could only evaluate performance on 2000 or less training data points. The GP hyperparameter optimization produced an output variance, which quickly approached zero as more data was added, leading to close to zero uncertainty everywhere, while the estimated probability stayed close to 0.5. For in distribution data, the DPMM gave probability estimates which were pretty noisy, as expected, but fluctuate around 0.5. Estimates for OOD data were approximately 0.5 and uncertainties were close to $0.5/\sqrt{3}$, in agreement with the chosen prior.

\section{Discussion and conclusion}
\label{sec:discussion}
To sum up, in the case of our two main datasets
\begin{itemize}
    \item the algorithms studied are all fairly well calibrated for in-distribution data where comparisons with the long-run frequency distribution make sense;
    \item the average uncertainty on the test set go down as we increase the number of training data points for all algorithms except evidential deep learning which did not exhibit this behavior under the hyperparameters we tested, and its uncertainties seem hard to interpret in the case of our data;
    \item the out-of-distribution average uncertainty as a function of number of training data points is high and goes down for the Gaussian process classifier and Dirichlet process mixture model, and is low or zero and constant for the fully connected neural network models with ReLU activation functions;
    \item for out-of-distribution data all fully connected neural network models with ReLU activation functions estimated extreme probabilities, while the nonparametric Bayesian models estimate probabilities around 0.5, with the Dirichlet process mixture model estimating slightly higher probabilities as the number of training data points increase, indicating some numerical instability or other error in implementation.
\end{itemize}
The DPMM is explicitly defined as having a $\text{Beta}(1,1)$ prior for class probability, and results support the qualitative correctness of the approximate posterior. For the neural networks and GP, the prior is implicitly defined, and so it is harder to evaluate the cohesiveness of the results. The approximate posterior distribution of the deep learning models behaves as if the probability mass is concentrated at 0 or 1. It is tempting to interpret this as an approximate posterior distribution with $\text{Beta}(0,0)$ priors for class probability. This interpretation suggests that changing the implicit prior for the deep learning models through modifying the loss function could change the observed behavior. However, it may not be a good idea to assign those kind of interpretations post-hoc, as we might put too much trust in a method without a principled theoretical foundation. Instead, we might consider if the approximations introduced by the approximate Bayesian inference methods (ensembles, MCD, variational approaches, etc.) are the cause of the deviation from desired results.

Although neural networks with rectified linear (ReLU) activation functions can in theory approximate any nonlinear function of the data locally, they eventually become simple linear functions of the input features if one moves in any direction far enough \citep{Hein2019}. This property of piecewise linear neural networks, combined with using the softmax function for normalization of the last layer of the network, gives rise to class probabilities that asymptotically approach extreme values of zero or one. 
Therefore the estimated extreme probabilities for out-of-distribution data may simply be a feature of using ReLU neural networks as the family of parameterized test distributions in approximate inference. Similar to more well known parametric methods such as linear regression we should not expect our model to be informative or correct if we extrapolate too far. Changing the architecture and activation functions to better reflect the kind of statistical model we seek is an obvious step to mitigate the problem of model misspecification. In this study we tried changing the activation function to one that is bounded to see if this would help the situation, but our results showed that this was not a robust solution although for some models it seemed to improve OOD uncertainties (see Appendix C).

It might also be the case that taking the shortcut of directly calculating the conditional class probabilities through supervised learning methods will never give us uncertainties that are dependent on the joint probability density. In that case, anomaly detection or other density estimation methods (such as the DPMM in this study) are obvious choices for estimates which warn about out-of-distribution data. Further studies, both empirical and theoretical, are needed to investigate these pathologies of neural networks and the effect of changing important elements like the activation function. There also exists many more algorithms for class probability estimation with uncertainties, such as different approximations of Bayesian neural networks, and looking into these is a topic for future research.

This study only investigates data which is sampled from a relatively simple two dimensional distribution and therefore does not represent the complexities present in real-world data. It is noteworthy that even for this relatively simple problem, many of the learning algorithms do not produce uncertainties which can be used by scientists to decide if they should trust the class probability estimates. Although the uncertainty estimates produced by the deep learning algorithms presented in this study were not especially useful for this type of data, they might work better for data with for example discrete or bounded distributions, or for data with statistically independent input features. We strongly recommended that scientists create toy data that more closely matches the mathematical properties of their own data of interest to see how the different machine learning algorithms can be expected to behave in their case.

Regardless of the mathematical reason for the observed behavior and the inherent limitation of empirical studies, we can take away some key lessons from the experiment. The studied methods for class probability and uncertainty estimates do not necessarily exhibit the behavior we expect or desire for out-of-distribution data. This may be partly due to a lack of understanding of what can reasonably be expected from the algorithms in question, like the fact that it is not possible to have unbiased frequency estimates which also extrapolate to a specific value in the $N\rightarrow0$ limit. It can also be because we do not yet have a clear understanding of the implicit statistical models created by our algorithms or the inherent limitations of the approximations we make in our inference algorithms to facilitate faster learning and compute intractable integrals. In any case, this study should serve as a reminder that uncertainty quantification for machine learning is still a largely unsolved problem. Sanity checks with intuitive toy-problems such as this one are a useful tool to explore the way these algorithms behave, and we hope this study can serve as an inspiration and a guide for future machine learning development work in the natural sciences.

\acks{We acknowledge support from the Research Council of Norway, grant no 314472, which is supporting the research of A.G.}

\bibliography{refs.bib}

\appendix
\section{Algorithms} \label{sec:A1}

In this section we describe the theoretical interpretation as well as any assumptions and computational details of the six algorithms used in the study.

\subsection{Neural networks with stochastic gradient descent}

A simple approximation of the integral in Equation \ref{eq:probability_MC} may be obtained by assuming that the only significant contribution to the integral comes from values of $\boldsymbol{\theta}$ close to the value $ \boldsymbol{\theta}_{\text{max}}$ which maximizes the posterior $p(\boldsymbol{\theta}|\mathcal{D_\text{train}})$. This approximation is appropriate if there is just one mode, which is close to the mean and most of the probability mass is centered around it. This is the case if we assume the posterior to be approximately Gaussian. The maximum a posteriori estimate may be found by any optimization method which can maximize the posterior with respect to $\boldsymbol{\theta}$. In the case of deep learning we use stochastic gradient descent derived optimization techniques for expectation maximization with regularization, which may be interpreted in a Bayesian way as sampling from this assumed Gaussian-like posterior distribution as defined by the NN architecture and the stochastic gradient descent algorithm and hyperparameters \citep{Duvenaud16}. 

All deep learning algorithms in the study were implemented in \texttt{pytorch}. All deep learning classifiers are simple fully connected neural networks with a \texttt{softmax}-function as the last normalizing layer unless otherwise specified. Details on the chosen architectures and hyperparameter settings are presented in Appendix B.

\subsection{Neural Network ensembles}

In neural network ensemble models we independently train $N_\text{E}$ neural networks to get a set of samples $\boldsymbol{\hat{\theta}_{\text{max}, i}}$ from the approximate posterior distribution and use these to compute the estimated probability and uncertainty using Equations \ref{eq:probability_MC} and \ref{eq:uncertainty_MC} as

\begin{equation*}
    \bar{Q}_\text{NNE}(\mathbf{c}|\mathbf{x}, \mathcal{D}_\text{train}) = \frac{1}{N_E}\sum_{i=1}^{N_E}Q(\mathbf{c}|\boldsymbol{\hat{\theta}_{\text{max}, i}},\mathbf{x}),
\end{equation*}
and
\begin{equation*}
    u_\text{NNE}^2(\mathbf{c}|\mathbf{x}, \mathcal{D}_\text{train}) = \frac{1}{N_E}\sum_{i=1}^{N_E}(Q(\mathbf{c}|\boldsymbol{\hat{\theta}_{\text{max}, i}},\mathbf{x}) - \bar{Q}_\text{NNE}(\mathbf{c}|\mathbf{x}, \mathcal{D}_\text{train})^2.
\end{equation*}
We set the number of ensembles to $N_\text{E}=20$ as in realistic cases it is usually not feasible to train many more ensembles than that. We tested varying $N_\text{E}$ between 5 and 50, and found that this made no significant difference to the results.

\subsection{Conflictual loss}
Conflictual loss (CL) is an algorithm proposed by \cite{Fellaji2024} which aims to give the ensemble uncertainties the property of being higher where there is little data and lower for network ensembles with fewer parameters. This is done by biasing an equal number of networks in the ensemble towards each class by adding a ``fake data"-term to the loss function 

\begin{equation} \label{eq:CL}
    L_{\text{CL}, i}(\mathcal{D}_N,\boldsymbol{\theta}) = \text{CE}(\mathcal{D}_N, \boldsymbol{\theta}) -\sum_{i=1}^{N} \sum_{j=1}^{k}\beta \log Q(c^b=1|\boldsymbol{\theta},\mathbf{x_i}) ).
\end{equation}
Here $i$ is the index of the network in the ensemble, $b$ is the index of the class that network is biased towards and $\beta$, the bias weight, is a hyperparameter which governs the strength of the bias. The cross entropy loss of N data points is given by
\begin{equation*}
    \text{CE}(\mathcal{D}_N, \boldsymbol{\theta}) = -\sum_{i=1}^{N} \sum_{j=1}^{k} c^j_i\log Q(c^{j}=1|\boldsymbol{\theta},\mathbf{x_i}).
\end{equation*}
The estimated probability and uncertainty is calculated as in NNE. This method can be seen as an attempt at indirectly changing the strength of the bias in Equation \ref{eq:bernoulli_mean}.

\subsection{Monte Carlo Dropout}

In deep learning using Monte Carlo Dropout (MCD) for uncertainty estimation \citep{Gal2016}, the ``dropping out" of weights commonly used during training to prevent overfitting, is applied at both training and inference phases. Specifically, during each forward pass through the network, random weights are set to zero according to a specified hyperparameter referred to as the dropout rate. MCD is related to ensemble learning. By performing multiple forward passes with different dropout masks, we generate a distribution of networks with associated predictions for a given input. Using a neural network with one infinitely wide hidden layer and MCD can theoretically be shown to be equivalent to a Gaussian Process, and therefore MCD is frequently referred to as approximate Bayesian inference. The estimated MCD probabilities and uncertainties presented in this study are the sample mean and standard deviations of 200 forward passes through the network. Changing the number of forward passes to 50 or 500 instead had no significant effect on our results.

\subsection{Evidential deep learning}

Evidential deep learning (EDL), as presented by \citep{Sensoy2018} which is based on the mathematical framework of subjective logic, replace the class probability point estimates of a neural network classifier trained on $k$ classes with a Dirichlet density distribution

\begin{equation*}
    D(\boldsymbol{\nu}|\boldsymbol{\alpha}) =
        \frac{1}{\text{B}( \boldsymbol{\alpha} )} \prod^k_{i=1} (\nu^i)^{\alpha^i-1},
\end{equation*}
where the components of $\nu$ belong to the standard $k-1$ simplex and $\text{B}(\boldsymbol{\alpha})$ is the multivariate beta function. $\alpha^j = e^j +1$ where $e^j$ is the ``evidence", in this case $e^j=\texttt{softplus}(f^j)=\log(1+\exp(f^j))$ of the logit outputs $\mathbf{f}(\mathbf{x}, \boldsymbol{\theta}) \in \mathbb{R}^k$ of the last non-softmax layer of the deep neural network.

The network is trained using the loss function 
\begin{align*} \label{eq:CDL}
    L_{\text{EDL}}(\mathcal{D}_N,\boldsymbol{\theta}) = &\sum_{i=1}^{N} \sum_{j=1}^k c^k_i(\psi(S(\mathbf{x}_i)-\psi(\alpha(\mathbf{x}_i)^j_i)) + \\&\lambda(t) \sum_{i=1}^{N} \text{KL}[D(\boldsymbol{\nu}_i|\tilde{\boldsymbol{\alpha}(\mathbf{x}_i)_i} || D(\boldsymbol{\nu}_i|<1,...,1>)],
\end{align*}
where $S_i=\sum_{k=1}^k\alpha_i^k$ and $\tilde{\boldsymbol{\alpha}}=\mathbf{c}_i + (1- \mathbf{c}_i)\cdot \boldsymbol{\alpha}(\mathbf{x}_i)$. This is the sum of the Bayes risk and the weighted KL-divergence between the estimated distribution and the uniform Dirichlet distribution. The annealing weight $\lambda(t)$ determines the strength of the second term, and is usually implemented as a linear function of the current epoch during training. In our case $\lambda(t) = \lambda_0t/200$ and $\lambda_0$ is a hyperparameter. The training algorithm aims to ensure that data points with no evidence are assigned a low belief mass, resulting in high uncertainty.

The estimated probabilities and uncertainties used in the study are the mean and standard deviation of the resulting distribution:

\begin{equation*}
    \bar{Q}_{EDL}(\mathbf{c}|\mathbf{x}, \mathcal{D}_\text{train}) = \mathbb{E}_D[\nu(\mathbf{c}|\mathbf{x})] = \frac{\boldsymbol{\alpha}}{\sum_{i=1}^K \alpha^i}
\end{equation*}

\begin{equation*}
    u_\text{EDL}^2(\mathbf{c}|\mathbf{x}, \mathcal{D}_\text{train}) = \text{Var}[\nu(\mathbf{c}|\mathbf{x})] = \frac{\mathbb{E}[\nu(\mathbf{c}|\mathbf{x})](1-\mathbb{E}[\nu(\mathbf{c}|\mathbf{x})])}{\sum_{i=1}^K \alpha^i+1}.
\end{equation*}
Our implementation of EDL is taken from the original paper's GitHub repository\footnote{\hyperlink{https://github.com/clabrugere/evidential-deeplearning/tree/main}{https://github.com/clabrugere/evidential-deeplearning/tree/main}}.

\subsection{Nonparametric Bayesian inference}

Nonparametric Bayesian inference algorithms provide principled methods of predictive inference for complex data. The connection between nonparametric Bayesian statistics and deep learning provides a bridge between the disciplines of statistics and machine learning and allows us to approach the latter with a well developed theoretical framework. We have included two nonparametric Bayesian inference algorithm in this study to show the similarities and differences to the approximate deep learning algorithms. 

\subsection{Gaussian Process}

A Gaussian Process (GP) is a Bayesian nonparametric approach to machine learning and closely related to deep learning \citep{Rasmussen2006}. In a GP model, we assume the relation $y = f(x) + \epsilon_y$ where $y$ is the variate we want to predict, $x$ are the rest of the variates, $f(x)$ is some function and $\epsilon_y$ is normally distributed noise. This makes $f(x) = y - \epsilon_y$ also normally distributed as $\mathcal{N}(\mu_f, \sigma_f)$, with $\mu_f$ and $\sigma_f$ given by the kernel function $\kappa(x, x_*)$ whose value is dependent on the relation between the training data points $x$ and the test point(s) $x_*$. In a binary classification task, the GP outputs the mean and variance of the logits $f_* =f(\mathbf{x_*})$ for each test point $x_*$ which we may transform to have the properties of a binomial probability distribution through the \texttt{sigmoid} function to perform logistic regression. 

If we use a kernel function which diverges for $x$ far from $\mu_x$, such as the radial basis function kernel, the estimated class probabilities converge towards $1/2$ far from the data and the estimated uncertainties also converge towards some value decided by the implicit prior given by the chosen kernel and hyperparameters. To keep the mathematical framework consistent with the rest of the analysis and bypass the difficulty of evaluating the intractable integral caused by the sigmoid function, probabilities and standard uncertainties are approximated using $T=1000$ Monte Carlo samples generated from $\mathcal{N}(\mu_{f*}, \sigma_{f*})$ \footnote{The difference between these probability estimates for our test set and other approximations of the class probabilities such as the approximation by David J.C. MacKay \citep{MacKay1992b} or the implementation used in Scikit taken from \citep{Williams1998} is less than 0.01.}:

\begin{equation*}
    \begin{aligned}
        \mathbb{E}_{f_*}[\texttt{sigmoid}(f_*)]= \int \texttt{sigmoid}(f_*)p(f_*|\mathcal{D}_N)df_*\\
        \approx \frac{1}{T}\sum_{t=1}^T \texttt{sigmoid}(f_{*,t})=\bar{Q}_\text{GP}(\mathbf{c}|\mathbf{x}, \mathcal{D}_N)
    \end{aligned}
\end{equation*}

\begin{equation*}
    u_{GP}(\mathbf{c}|\mathbf{x}, \mathcal{D}_N)^2 = \frac{1}{T}\sum_{t=1}^T (\texttt{sigmoid}(f_{*,t})- \bar{Q}_\text{GP}(\mathbf{c}|\mathbf{x}, \mathcal{D}_N))^2
\end{equation*}

with $p(f_*|\mathcal{D}_N)=\mathcal{N}(\mu_{f*}, \sigma_{f*})$.

We used a modified version of the \texttt{GaussianProcessClassifier} class in \texttt{scikit learn} \citep{scikit-learn} to extract the mean and standard deviation of the latent function posterior distribution. A radial basis function kernel with initial output variance 2 and length scale 8 was used. These parameters are automatically optimized and are presented in Table \ref{tab:GP}. The use of a radial basis kernel with optimized parameters expresses a prior belief that we are looking for smooth functions of limited amplitude.

\begin{table}
    \caption{Optimal hyperparameters for the Radial Basis Function kernel of the Gaussian Process for increasing number of training data points.}

    \begin{subtable}{.5\linewidth}
        \centering
        \caption{Dataset A}
        \begin{tabular}{l|c|c }
              $N_\text{train}$ & $l$ & $\sigma_o$ \\
              \hline
             250 & 5.45 & 1.97 \\
             500 & 8.97 & 1.47 \\
             1000 & 8.15 & 1.38 \\
             2000 & 7.58 & 1.34 \\
             3000 & 7.4 & 1.26 \\
             5000 & 7.24 & 1.14 \\
             10000 & 7.21 & 1.09 \\
             \hline
        \end{tabular}
    \end{subtable}%
    \begin{subtable}{.5\linewidth}
        \centering
        \caption{Dataset B}
        \begin{tabular}{l|c|c }
              $N_\text{train}$ & $l$ & $\sigma_o$ \\
              \hline
             250 & 8.65 & 1.87 \\
             500 & 9.43 & 1.84\\
             1000 & 9.09 & 1.79\\
             2000 & 5.94 & 1.71\\
             3000 & 5.43	& 1.67\\
             5000 & 5.85 & 1.62\\
             10000 & 6.57 & 1.63\\
             \hline
        \end{tabular}
    \end{subtable}%
    \label{tab:GP}
\end{table}

\subsection{Dirichlet Process Mixture Model}
The last model in the study is a Dirichlet process mixture model (DPMM) of product kernels. The Dirichlet Process is a stochastic process which generalizes the Dirichlet distribution, the conjugate prior for a discrete nominal variate with a fixed number of categories, into the prior for infinitely many categories, making the mixture model nonparametric \citep{Li2019}. Unlike the GP, this framework relaxes the global Gaussian process assumptions and can model multi-modal or non-smooth structures. The algorithm is also the only one in the study that approximates the full joint distribution $p(\mathbf{c}, \mathbf{x} | \mathcal{D})$. This makes the data model powerful but the calculations become more difficult and require extensive resource demanding Markov-Chain Monte Carlo simulations. The predictive posterior probability for a new data point is given by

\begin{equation*}
    p(\mathbf{d}|\mathcal{D_\text{train}}) = \int Q(\mathbf{d}|\mathbf{w}, \boldsymbol{\alpha})dp(\mathbf{w},\boldsymbol{\alpha}|\mathcal{D_\text{train}})
    = \int \sum_{i=1}^M  w_i  \prod_{j=1}^{n+1}A_j(\mathbf{d}_j,\boldsymbol{\alpha}_{i,j}) dp(\mathbf{w},\boldsymbol{\alpha}|\mathcal{D_\text{train}}),
\end{equation*}

where $\mathbf{w}$ are the mixture weights, $A_j$ are feature specific parametric distributions with mixture-dependent parameters $\boldsymbol{\alpha}_{i,j}$, $n$ is the dimensions of $\mathbf{x}$ and $M$ are the number of non-zero mixture components. Knowing the joint distribution lets us calculate any quantity of interest such as conditional or marginal distributions using the normal rules of probability calculus. The specific parametric family of distributions, $A_j$, that are attributed to each variate and their associated prior distributions influence the modeling and computational performance and should be chosen carefully, although the available choices are limited by computational and numerical restraints \citep{gorur2010}.

For our study we used the implementation in v0.2.1 of the R-package \texttt{inferno} \citep{inferno}\footnote{\hyperlink{https://github.com/pglpm/inferno/}{https://github.com/pglpm/inferno/}} based on the mathematical framework of \citep{Dunson2011, Ishwaran2002},  which is a flexible framework for regression and classification on manifolds. The prior distribution and hyperparameter choices are kept default as described in the accompanying documentation \citep{inferno_paper}. For the class variate, we used a mixture of Bernoulli distributions with a $\text{Beta}(1,1)$ prior, and for the input features we used a mixture of axis-aligned Gaussian distributions with prior hyperparameters automatically set by the data. Changing the value of the Beta hyperparameter $c$ corresponds to changing the strength of our ``noninformative" belief of equal probability in each class. 1200 samples of $\mathbf{w}, \boldsymbol{\alpha}$ for the approximation of the joint posterior distribution were collected using Gibbs sampling and 10 Markov-Chain Monte Carlo (MCMC) chains with a burn-in of 3600 samples and early stopping \citep{Gong2016}. Issues related to MCMC convergence and truncation can cause numerical artifacts in these kind of models, so it is important to inspect both the MCMC diagnostics results and the inference results carefully. We approximate the expected value and standard uncertainty of the conditional class probability by the sample mean and standard deviation of the conditional class probabilities calculated from the Monte Carlo samples.

\begin{equation*}
    \bar{Q}_\text{DPMM}(\mathbf{c}|\mathbf{x}, \mathcal{D}_N) = \frac{1}{T}\sum_{t=1}^T Q(\mathbf{c}|\mathbf{x}, \mathbf{w}_t,\boldsymbol{\alpha}_{t}) 
\end{equation*}

\begin{equation*}
    u_\text{DPMM}^2(\mathbf{c}|\mathbf{x}, \mathcal{D}_N) = \frac{1}{T}\sum_{t=1}^T  (Q(\mathbf{c}|\mathbf{x}, \mathbf{w}_t,\boldsymbol{\alpha}_{t})  - \bar{Q}_\text{DPMM}(\mathbf{c}|\mathbf{x}, \mathcal{D}_N))^2
\end{equation*}

\section{Neural network hyperparameters} \label{sec:A2}

\begin{figure}
    \begin{subfigure}[b]{0.45\textwidth}
        \includegraphics[width=\textwidth]{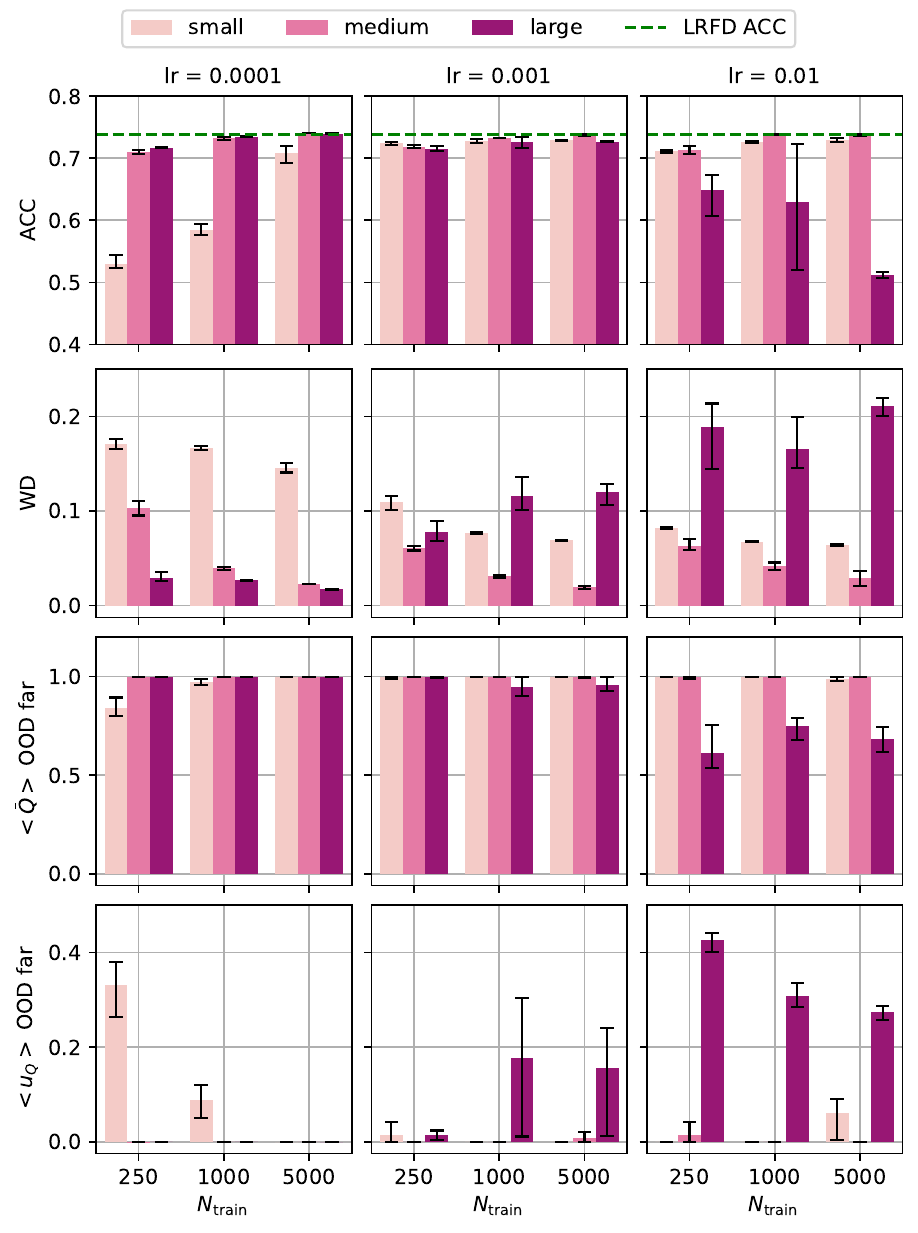}
        \caption{Dataset A}
    \end{subfigure}
    \hfill
    \begin{subfigure}[b]{0.45\textwidth}
        \includegraphics[width=\textwidth]{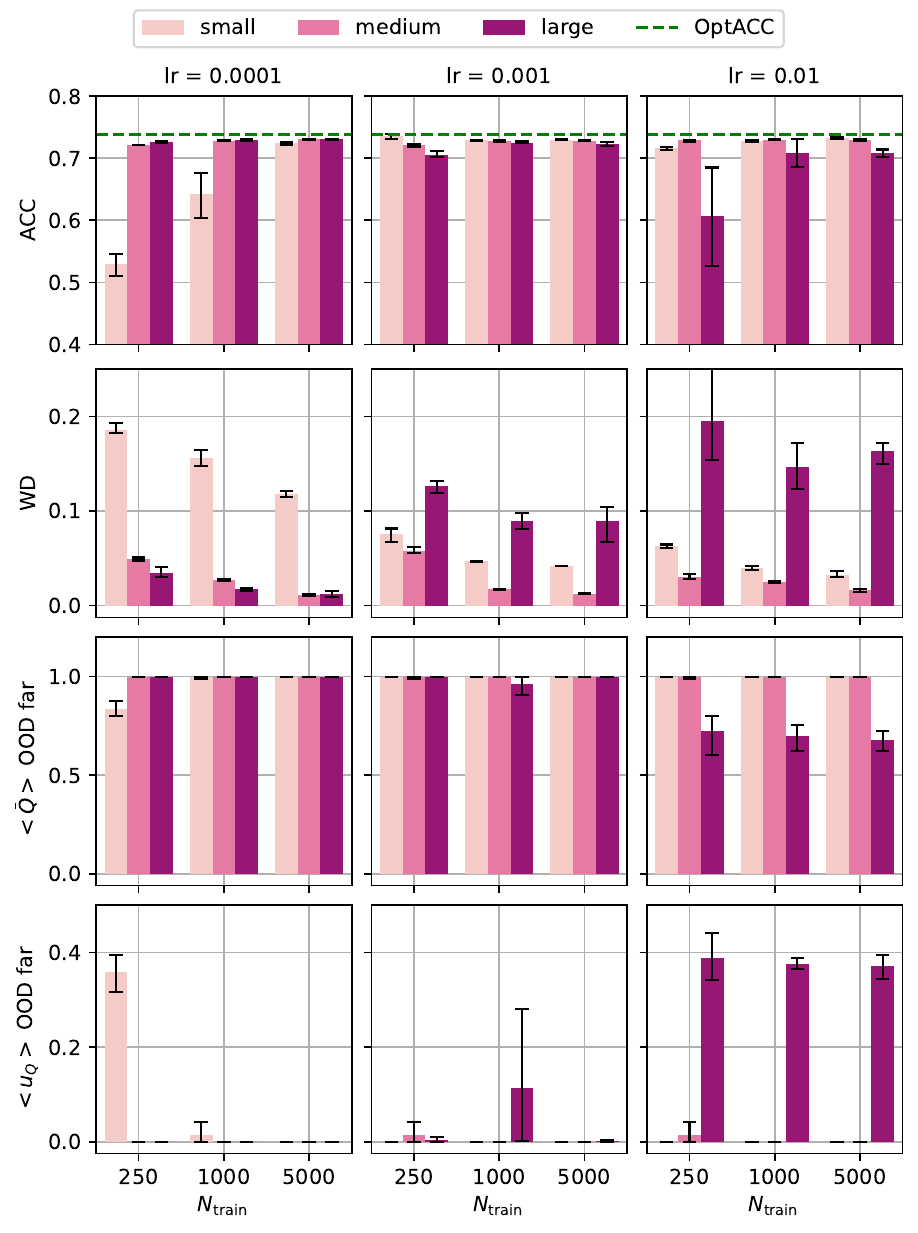}
        \caption{Dataset B}
    \end{subfigure}
    
    \caption{Results for different hyperparameter settings for neural network ensemble (NNE). The columns contain the results for learning rates 0.0001, 0.001 and 0.01. The color of the bars indicates the depth and width of the hidden layers of the network. Small, medium and large has 1, 3, and 8 hidden layers with 20, 200 and 2000 nodes respectively. The top row plots shows the accuracy on the validation set, with the green dashed line indicates the LRFD accuracy for that validation set. The second row plots shows the Wasserstein distance (WD) of the validation set. The two last rows of plots show the average estimated class probability and uncertainty for out-of-distribution test points. The error bars indicate the 2.5-97.5 percentiles.}
    \label{fig:gridsearch_NNE}
\end{figure}

\begin{figure}
    \begin{subfigure}[b]{0.45\textwidth}
        \includegraphics[width=\textwidth]{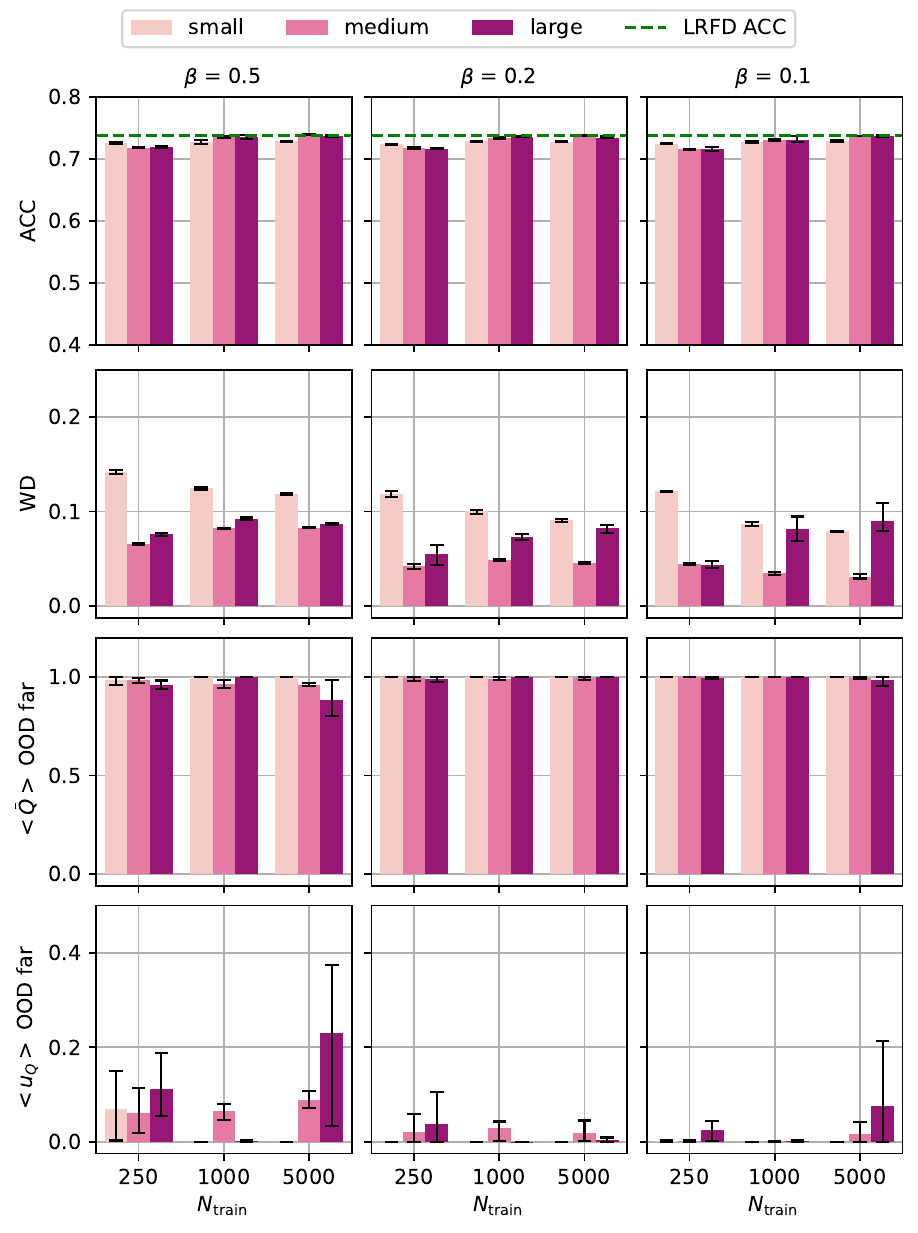}
        \caption{Dataset A}
    \end{subfigure}
    \hfill
    \begin{subfigure}[b]{0.45\textwidth}
        \includegraphics[width=\textwidth]{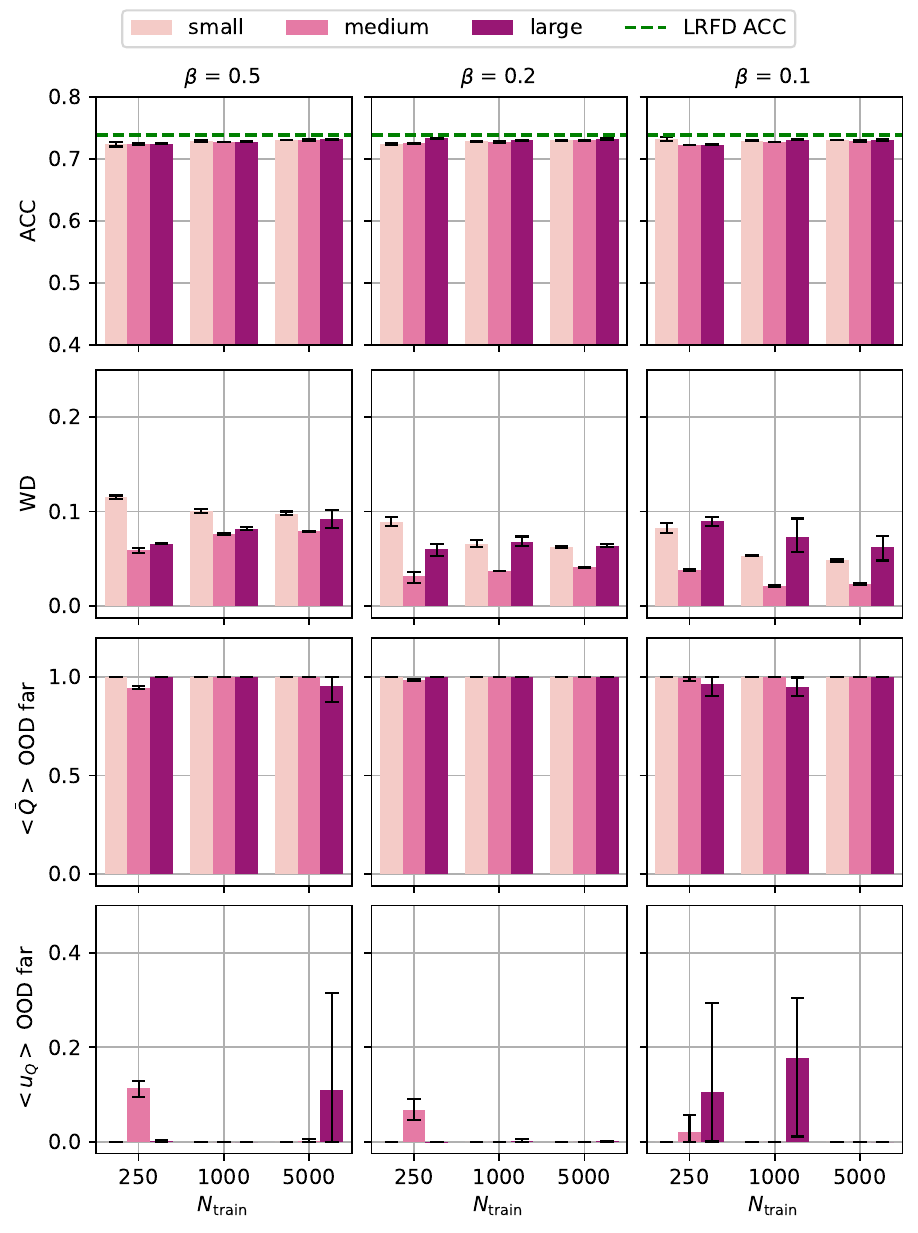}
        \caption{Dataset B}
    \end{subfigure}
    
    \caption{Results for different hyperparameter settings for neural network ensemble with conflictual loss (CL). Same as for Figure \ref{fig:gridsearch_NNE} except all the models are trained with learning rate 0.001 and the columns show results for different values of the bias weight $\beta$.}
    \label{fig:gridsearch_CL}
\end{figure}

\begin{figure}
    \begin{subfigure}[b]{0.45\textwidth}
        \includegraphics[width=\textwidth]{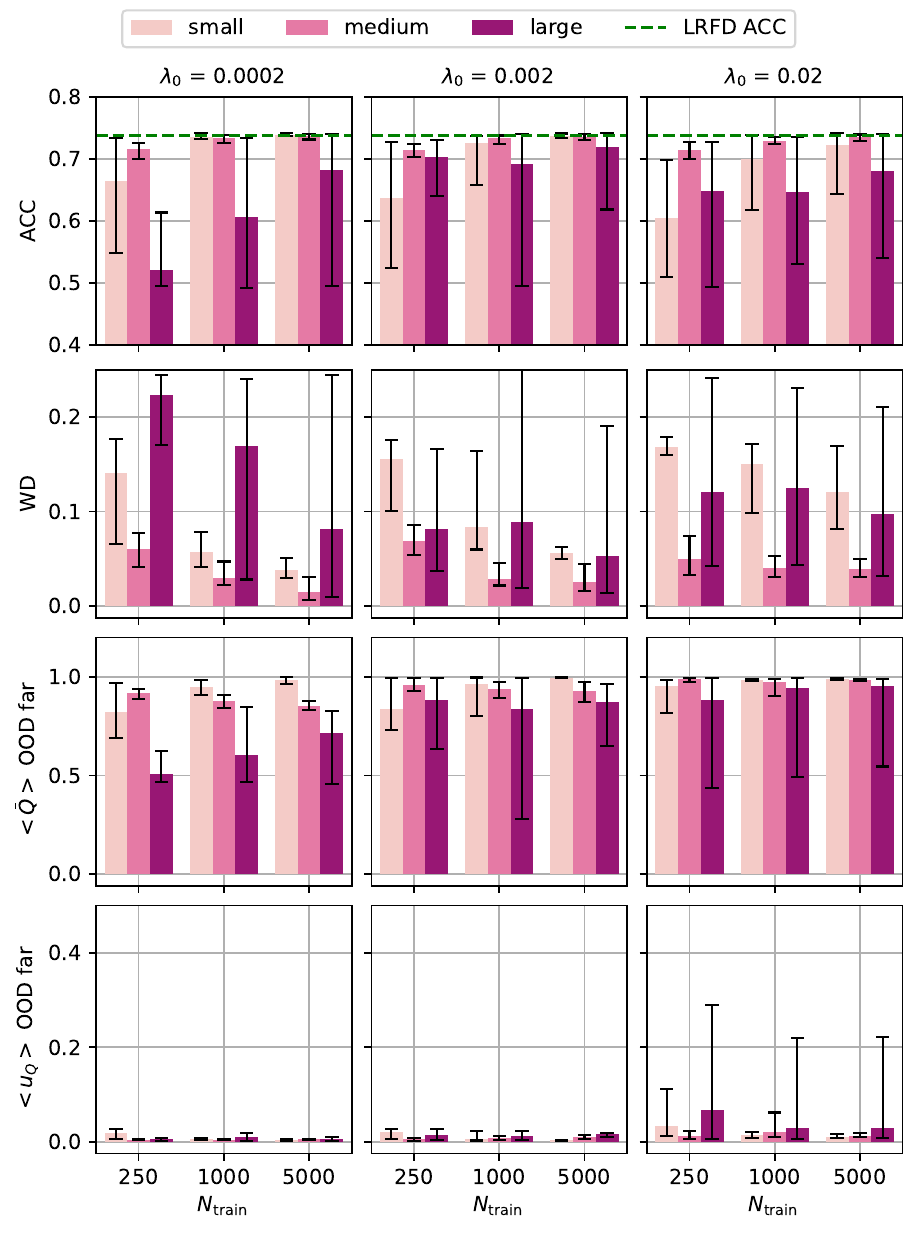}
        \caption{Dataset A}
    \end{subfigure}
    \hfill
    \begin{subfigure}[b]{0.45\textwidth}
        \includegraphics[width=\textwidth]{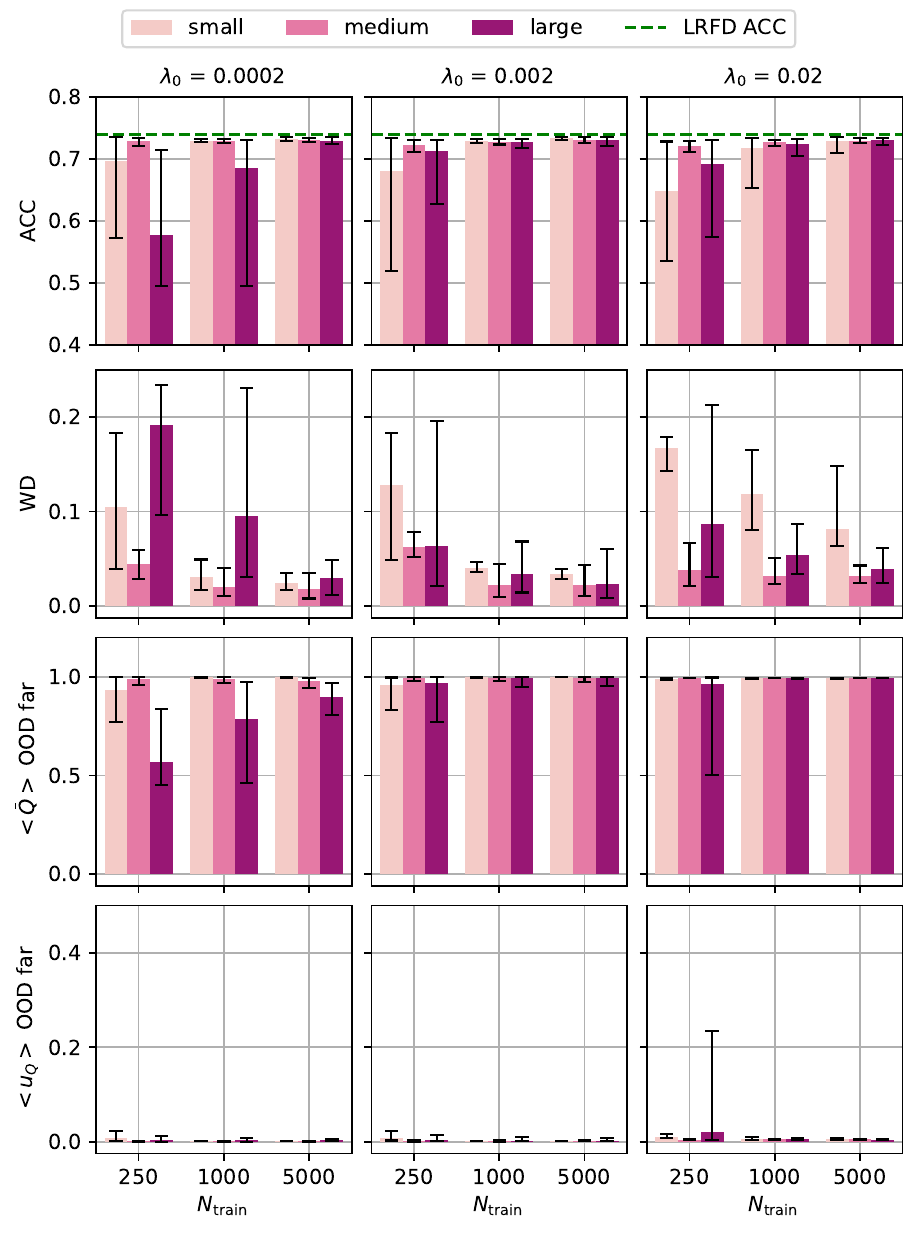}
        \caption{Dataset B}
    \end{subfigure}
    
    \caption{Results for different hyperparameter settings for evidential deep learning (EDL). Same as for Figure \ref{fig:gridsearch_NNE} except all the models are trained with learning rate 0.0001 and the columns show results for different values of the annealing coefficient $\alpha$.}
    \label{fig:gridsearch_EDL}
\end{figure}

\begin{figure}
    \begin{subfigure}[b]{0.45\textwidth}
        \includegraphics[width=\textwidth]{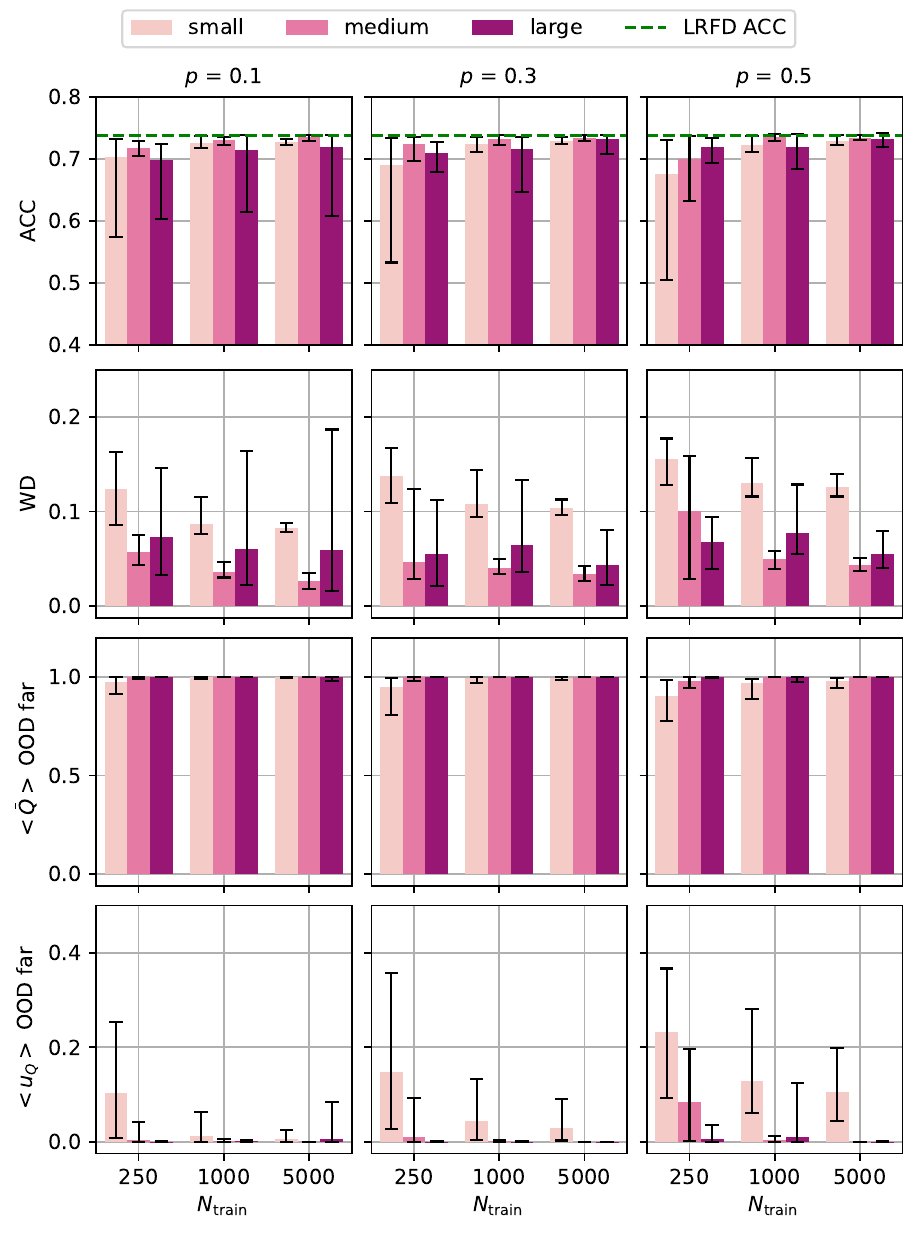}
        \caption{Dataset A}
    \end{subfigure}
    \hfill
    \begin{subfigure}[b]{0.45\textwidth}
        \includegraphics[width=\textwidth]{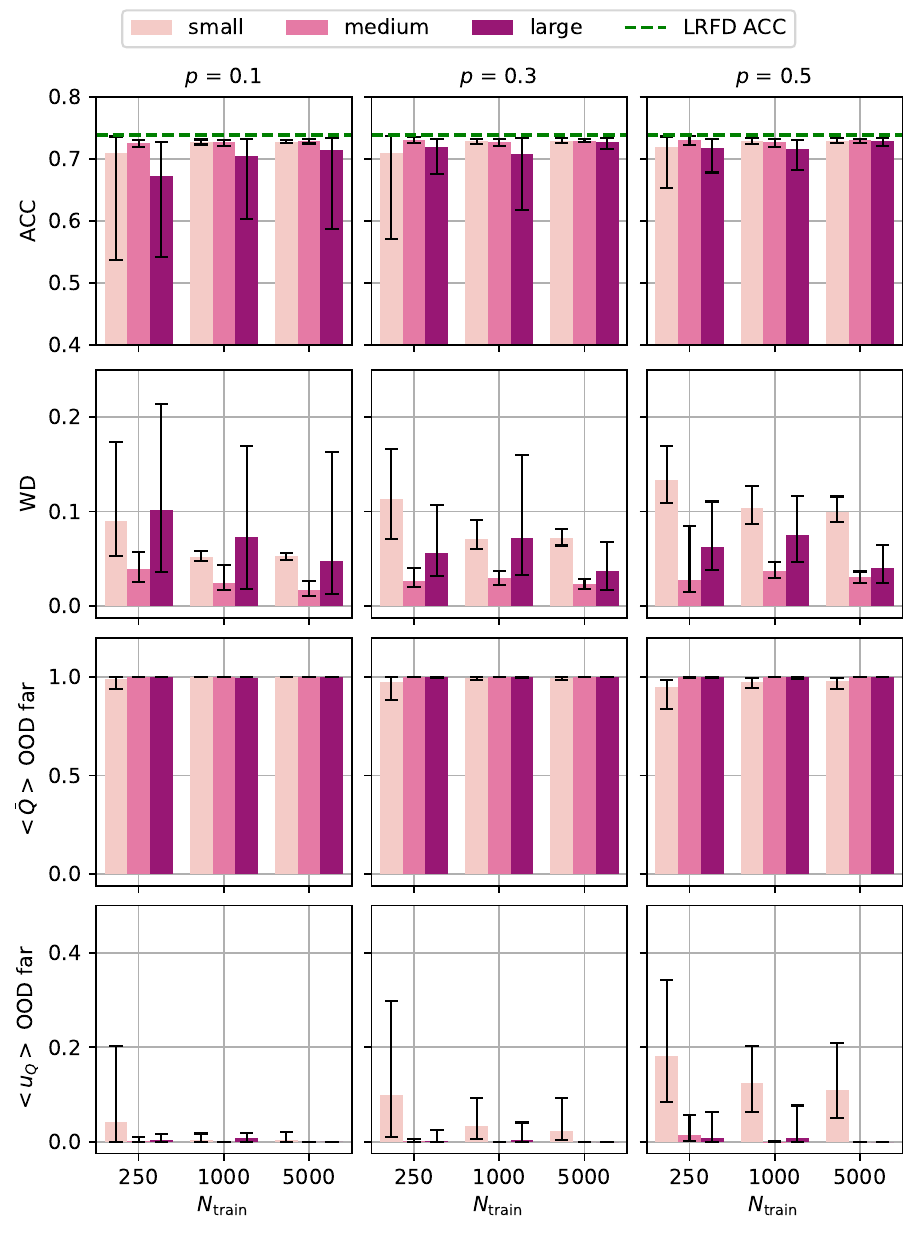}
        \caption{Dataset B}
    \end{subfigure}
    
    \caption{Results for different hyperparameter settings for neural network with Monte Carlo dropout. Same as for Figure \ref{fig:gridsearch_NNE} except all the models are trained with learning rate 0.001 and the columns show results for different values of the drop out rate $p$.}
    \label{fig:gridsearch_MCD}
\end{figure}

The neural networks we have used in the main study are all fully connected neural networks with ReLU activation functions. They are trained using stochastic gradient descent with the AdamW modification of the Adam optimizer \citep{Kingma2014}. The models are trained on batches of $[128, 128, 256, 256, 1024, 1024, 2048]$ samples for $N_\text{train} = [250, 500, 1000, 2000, 3000, 5000, 10000]$. The models were each trained for a maximum of 200 epochs (substantially more than needed) with early stopping if the loss did not decrease within 20 epochs. The early stopping criteria, as well as the way the hyperparameters are chosen as described below, biases the results towards networks with good global calibration properties and does not take OOD performance into account.

A grid search was run over hyperparameter settings where we investigated the influence of changing learning rate (0.01, 0.001 and 0.0001), weight decay (0.1, 0.01 and 0.001) and network size (1, 3, 8 hidden layer(s) with 20, 200, 2000 nodes respectively) as well as the bias weight for CL (0.1, 0.2, 0.5), the annealing coefficient for EDL (0.0002/200, 0.0002/200, 0.02/200) and the drop out rate for MCD (0.1, 0.3, 0.5). For the ensembles, one ensemble with 20 networks was trained for each of the 81 different combinations of hyperparameters, while for the single network EDL and MCD algorithms 20 separate networks were trained for each combination. Weight decay was found to not have any major influence on our results, so for the study, weight decay was set to 0.01 for all algorithms. 

The accuracy and Wasserstein distance (WD) of the validation set, the average OOD-far estimated class probability and the average OOD-far uncertainty was calculated and are presented in Figures \ref{fig:gridsearch_NNE}-\ref{fig:gridsearch_MCD} as bar plots with error bars indicating the 2.5-97.5 percentiles over different runs, including the runs with varying weight decay. The $x$-axis of the plots show the results for different number of training points, while the color of the bar indicates the network size. The accuracy is a good measure of whether the network is properly trained. For dataset A, the optimal accuracy of the validation set, if the long run frequency was known, is 73.82\% and for dataset B the accuracy is 73.88\%. Based on the results, any networks with less than 70\% accuracy even with just 250 data points should be considered very poorly fitted. 

The WD is a global measure of calibration, and for the validation set it should not be strongly influenced by any bias in the tails of the distribution. It is expected to go down with increasing number of training data, and hyperparameter settings which do not give results that follow this trend are considered to be poor choices. The OOD estimated probability and uncertainty is also shown to see if the choice of hyperparameter would influence the results of the study. Some hyperparameter combinations do produce high uncertainties and more conservative estimated class probabilities, seemingly in conflict with our study results, but these tend to have lower accuracy scores and higher WD scores. The rest of the OOD results confirm our conclusion that the NN-based estimated probabilities and uncertainties for our dataset approach extreme probabilities and zero uncertainty.

Learning rate had a significant effect on results, which can be seen in Figure \ref{fig:gridsearch_NNE} where the results of varying learning rate and network size as well as the number of training points for the NNE algorithm is presented. Small networks need a larger learning rate, while larger networks need a smaller learning rate. For the study the medium size neural network with 3 hidden layers with 200 nodes each and a learning rate for 0.001 for all deep learning algorithms was used.

The results of varying the bias weight of the CL algorithm are presented in Figure \ref{fig:gridsearch_CL}. Setting the bias weight to 0.1 produced the lowest WD-scores, so this setting was used in the study. Some hyperparameter settings produce high OOD uncertainties sporadically, but these are highly variable and so could not be used as a robust anomaly detection method.

The EDL algorithm performs best across network size for annealing coefficient $\alpha=0.002$. But for the ideal medium network size, it is unimportant. Larger and smaller networks show high variability in accuracy and mean WD. For dataset A, the mean estimated probability for OOD data is close to 1, but not always equal to 1. For dataset B, the estimated OOD probabilities are more extreme. Despite this, the OOD uncertainty is in most cases close to 0, even for several poorly trained networks.

For the MCD algorithm, the drop out rate does not seem to affect results much at all except for the smallest networks which also performed worse, as shown in Figure \ref{fig:gridsearch_MCD}. In those cases, a higher drop out rate led to lower accuracies and higher WD-scores. In the study we have used a drop out rate of 0.3. Some hyperparameter settings, mainly those with small networks produce high OOD uncertainties sporadically, but these are also the ones that perform the worst on the validation set.

We also checked the effect of different hyperparameter choices for the DPMM algorithm. We tested that the estimated OOD-far uncertainties were approximately equal to the predictions of equation \ref{eq:bernoulli_var} for $c=[0.1, 0.5, 1, 2]$. Lower values for c give a slower convergence (as a function of $r$) towards the theory predictions and vice versa. All values for $c$ showed an increase in the predicted class 1 probability estimates towards 1 for larger values of $N_\text{train}$. Class 1 is more dominant in the tails of the in-distribution data and may therefore be assigned higher probabilities that should in principle be balanced by the rest of the components in the mixture, but in practice are not. Inspections of the trace plots and diagnostics do not indicate that these results are due to MCMC convergence issues.

In summary, the grid search corroborates our main findings and provide interesting information on the effect of the choice of different indirect priors through the hyperparameter settings.

\section{Changing the activation function} \label{sec:A3}

\begin{figure}
    \centering
    \begin{subfigure}[b]{0.8\textwidth}
        \includegraphics[width=\textwidth]{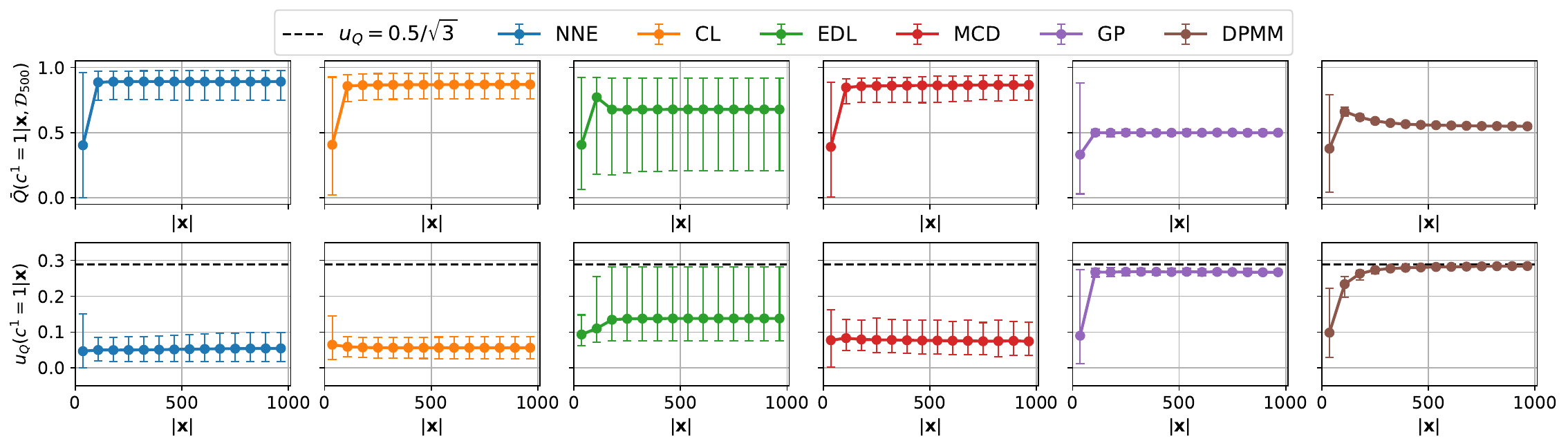}
        \includegraphics[width=\textwidth]{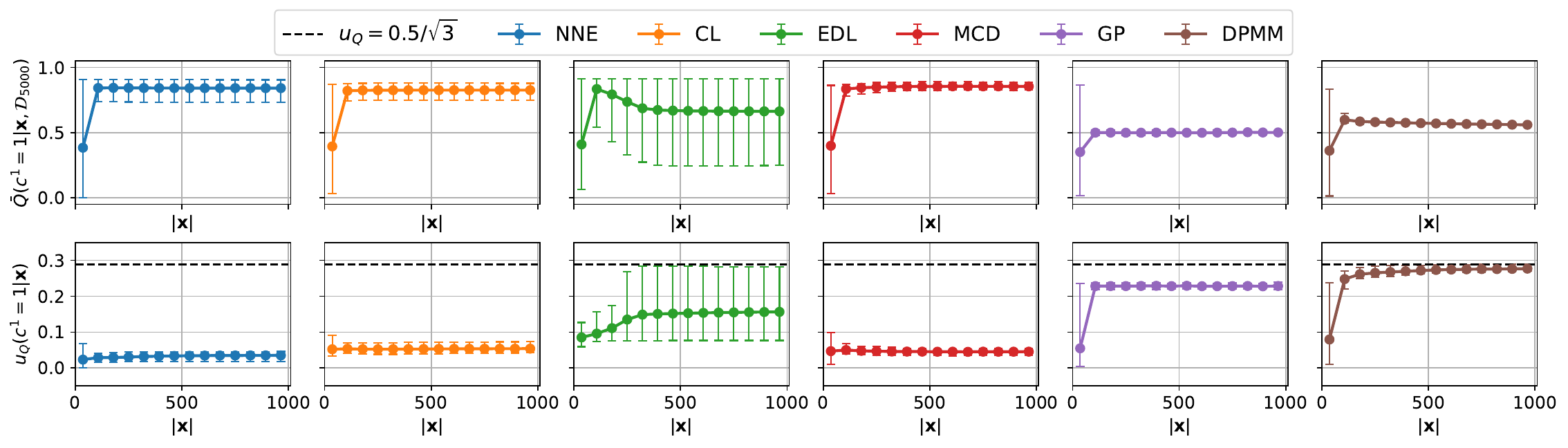}
        \caption{Dataset A.}
    \end{subfigure}
    \begin{subfigure}[b]{0.8\textwidth}
        \includegraphics[width=\textwidth]{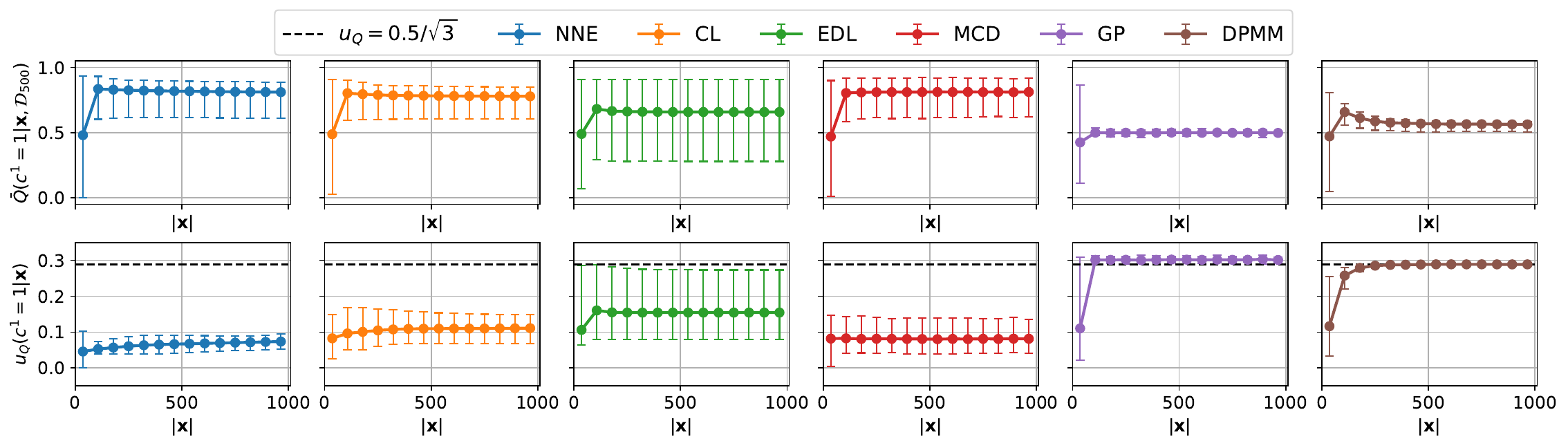}
        \includegraphics[width=\textwidth]{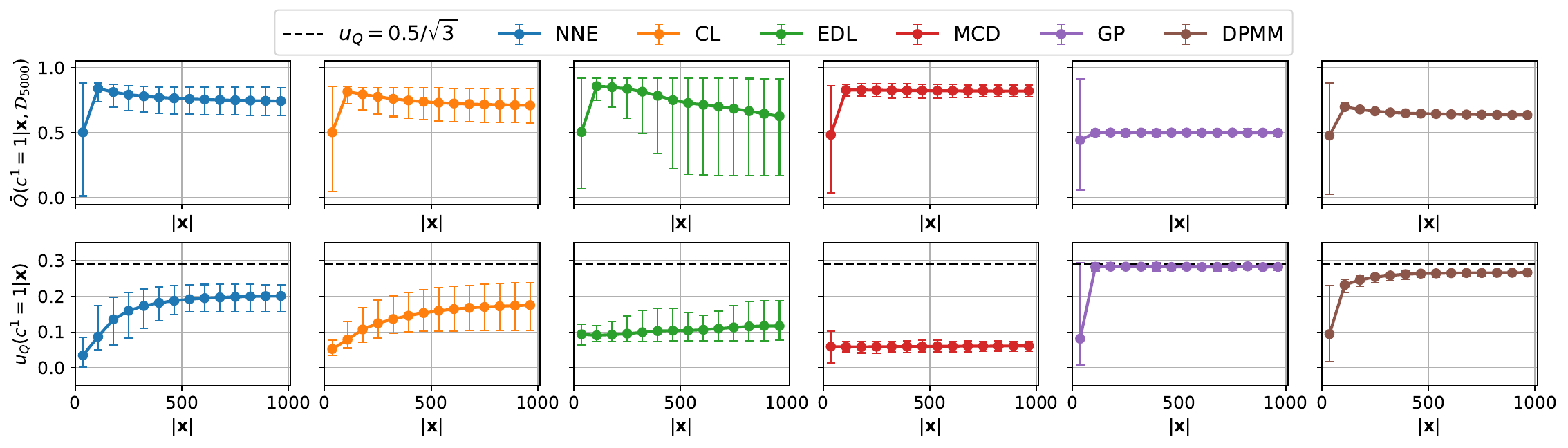}
        \caption{Dataset B.}
    \end{subfigure}
    \caption{Estimated probabilities (top row) and uncertainties (bottom row) for class 1 for the different algorithms for the two datsets as a function of radius $|\mathbf{x}|$. The error bars indicate the entire spread of the data over polar angle $\phi$, while the markers indicate the sample average.  The dashed black line in the plots in the bottom rows indicates the standard deviation of the predictive posterior of a binomial model with $N=0$ and prior $\text{Beta}(1,1)$.}

    \label{fig:extrapolation_tanh}
\end{figure}

\begin{figure}
    \centering
    \includegraphics[width=0.8\linewidth]{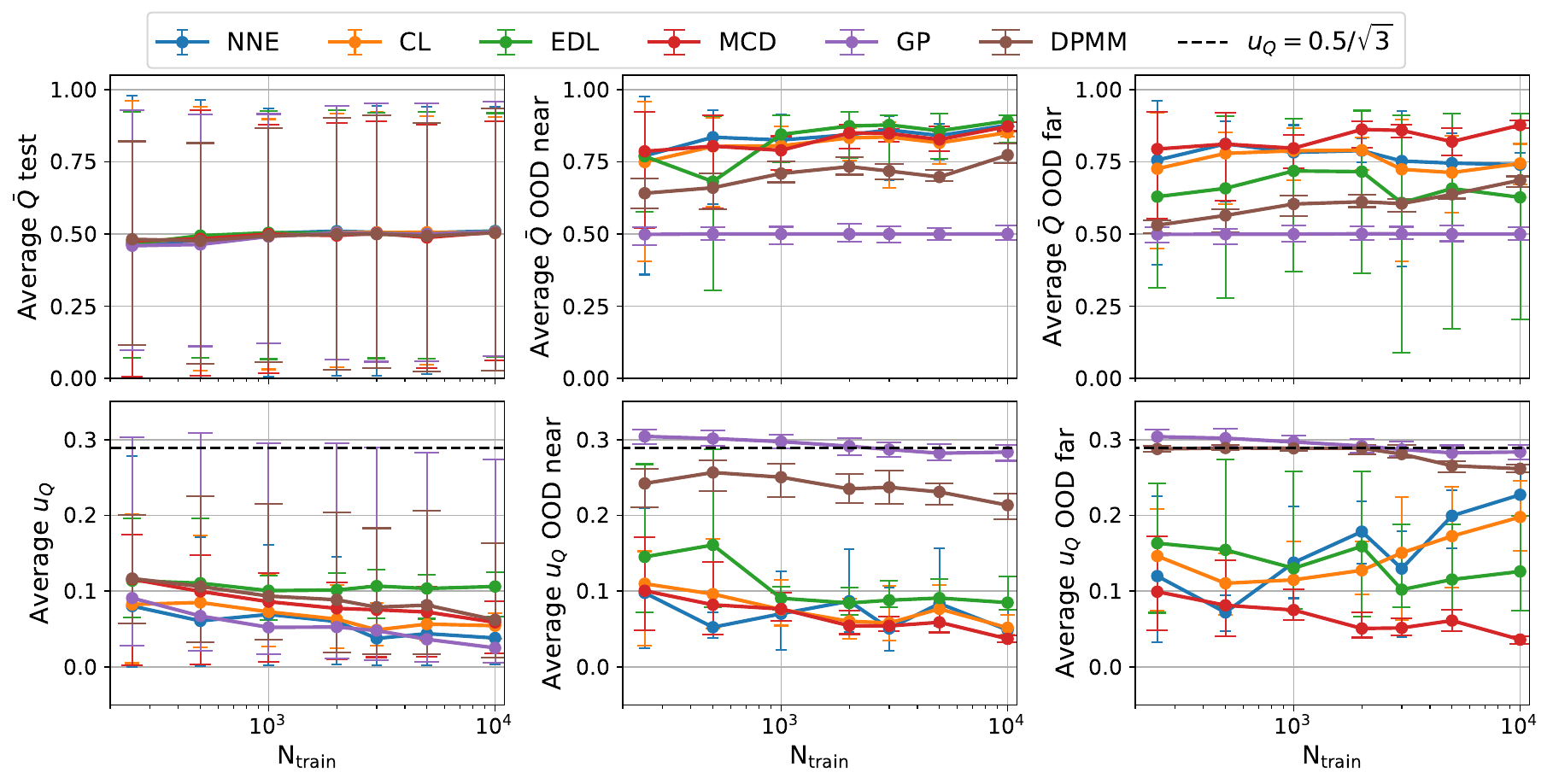}
    \caption{Top: Average estimated out-of-distribution probabilities (top rows) and uncertainties (bottom rows) as a function of $N_\text{train}$ for the test set (left), the OOD-near test set (middle) and the OOD-far test set (right) with the deep learning models using the $\text{tanh}$-activation function. The error bars indicate the entire spread of the data over polar angle $\phi$. The colors are similar to Figure \ref{fig:metrics}. The dashed black line in the plots in the bottom rows indicates the standard deviation of the predictive posterior of a binomial model with $N=0$ and prior $\text{Beta}(1,1)$.}
    \label{fig:uncertainties_B_tanh}
\end{figure}

We also tried changing the activation function in the neural networks to avoid the known pathologies of \texttt{ReLU} neural networks. Using the \texttt{tanh} activation function gave results that were qualitatively different from the previous results. Figure \ref{fig:extrapolation_tanh} shows the average and spread in class probability and uncertainty estimates over $\phi$ as a function of $r$ for values of $r$ ranging from 0 to 1000. In this case, the estimated probabilities stabilize around 0.8 instead of 1 and uncertainties are not zero. For dataset A there is no distinct increase in uncertainty for OOD data for NNE, CL and MCD, but for EDL there is an observed increase. However, the estimates vary wildly across $\phi$ and would not be reliable in a practical setting. For dataset B we see a profound increase in uncertainties as a function of increasing $r$ for NNE and CL for 5000 training data points. The trend holds for $N_\text{train}\geq 1000$ as can be seen in Figure \ref{fig:uncertainties_B_tanh}. The estimates still do not give class probabilities of 0.5, and uncertainties are much lower than DPMM and GP for the OOD-near test points. EDL estimates behave very unpredictably for this dataset while MCD has similar behavior as for dataset A.

\section{Reliability diagrams} \label{sec:A4}
\begin{figure}
    \centering
    \begin{subfigure}[b]{0.8\textwidth}
        \includegraphics[width=\textwidth]{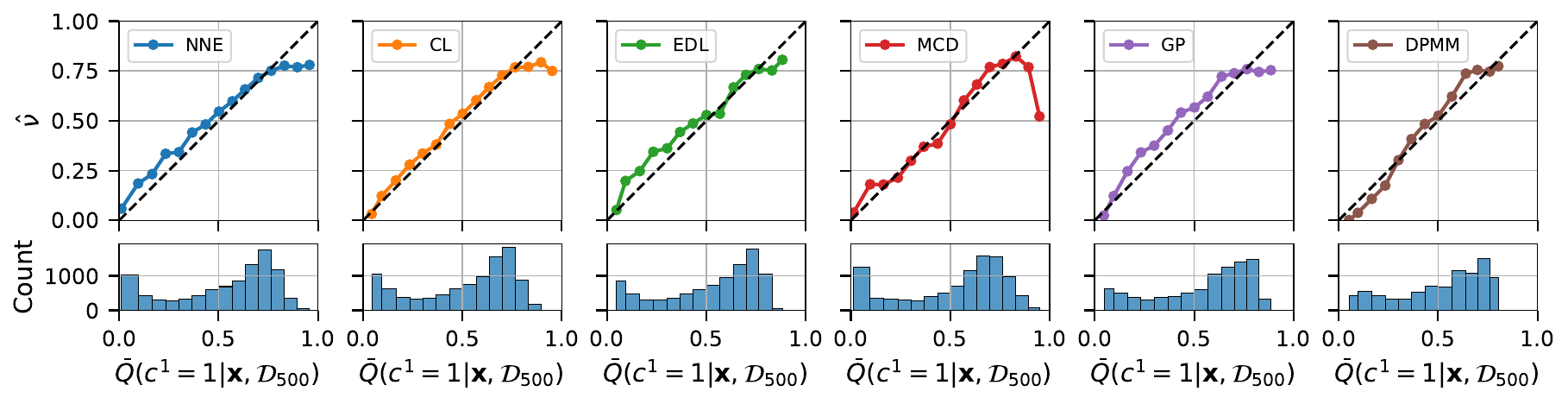}
        \includegraphics[width=\textwidth]{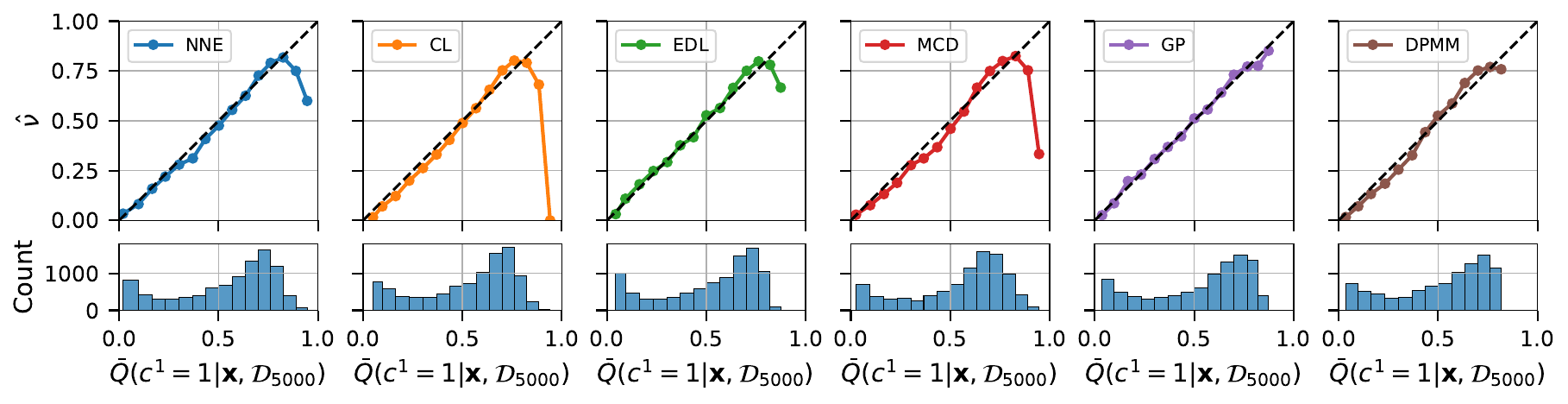}
        \caption{Dataset A}
    \end{subfigure}
    \hfill
    \begin{subfigure}[b]{0.8\textwidth}
        \includegraphics[width=\textwidth]{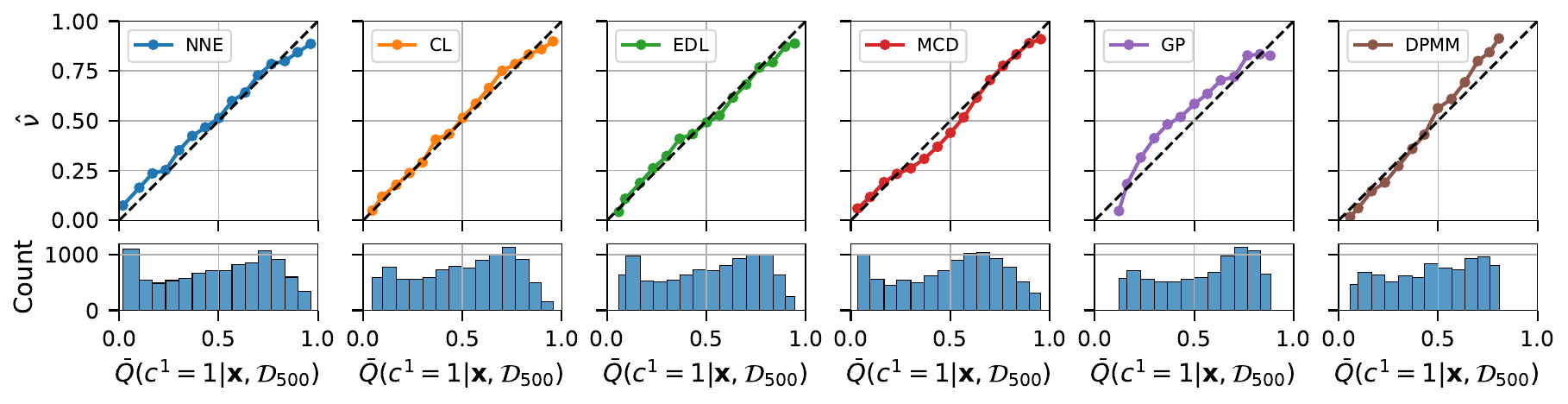}
        \includegraphics[width=\textwidth]{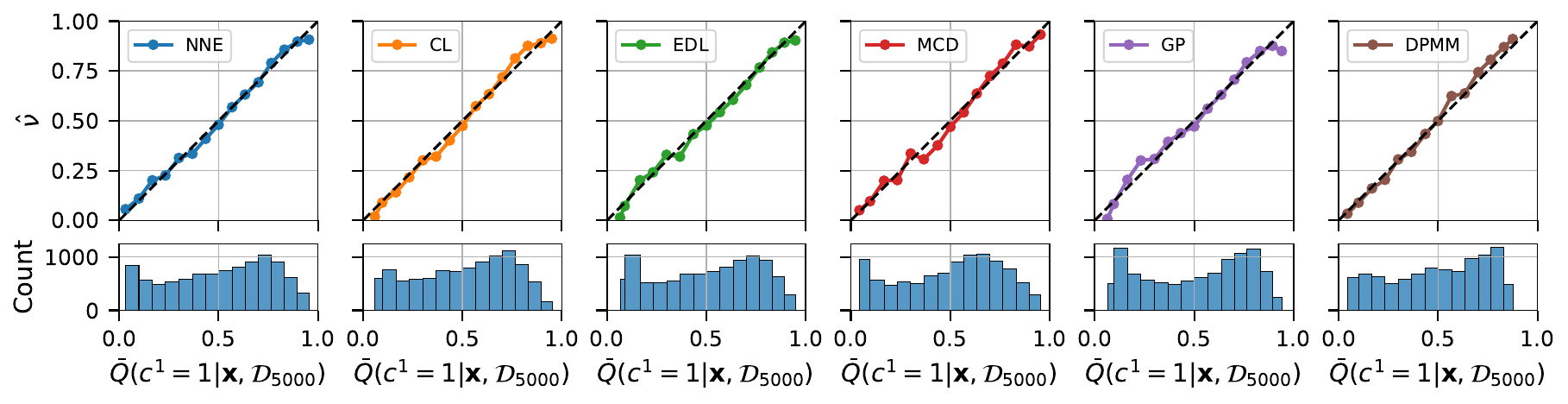}
        \caption{Dataset B}
    \end{subfigure}
    
    \caption{Reliability diagrams calculated using the test datasets over 15 uniform bins in $[0,1]$ for the two datasets. The the x-axis indicates the estimated class probability of class 1, while the y-axis of the diagram indicates the empirical frequency of class 1 of the bin. The top plots for each dataset are the results for $N_\text{train}=500$ and the bottom for $N_\text{train}=5000$. Under each reliability plot is a histogram showing the number of data points in the test set with estimated class probabilities belonging to that bin.}
    \label{fig:reliability_diagrams}
\end{figure}

Reliability diagrams are a common tool for evaluating calibration of class probabilities. The reliability diagrams for a selection of the trained models are presented in Figure \ref{fig:reliability_diagrams}. They were calculated using 15 uniform bins over $[0,1]$. The plots quantitatively support the claim that all models are well calibrated outside the tail of the distributions. For the bins close to 1, it can be seen that the deep learning models are overconfident with respect to the calculated empirical frequency distribution for dataset A, but not for dataset B.

\section{2D distributions} \label{sec:A5}
\begin{figure}
    \centering
    \includegraphics[width=\textwidth]{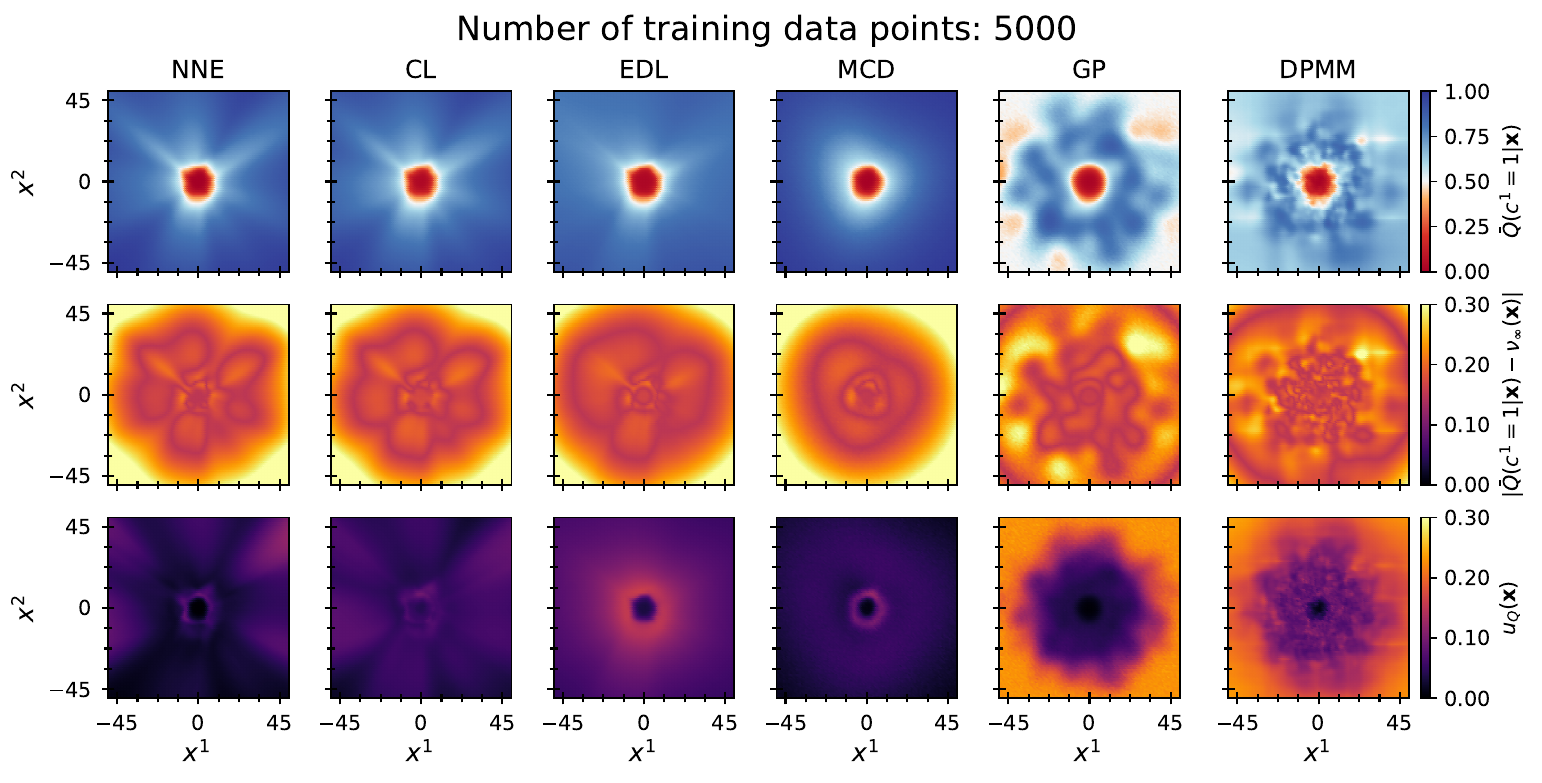}
    \includegraphics[width=\textwidth]{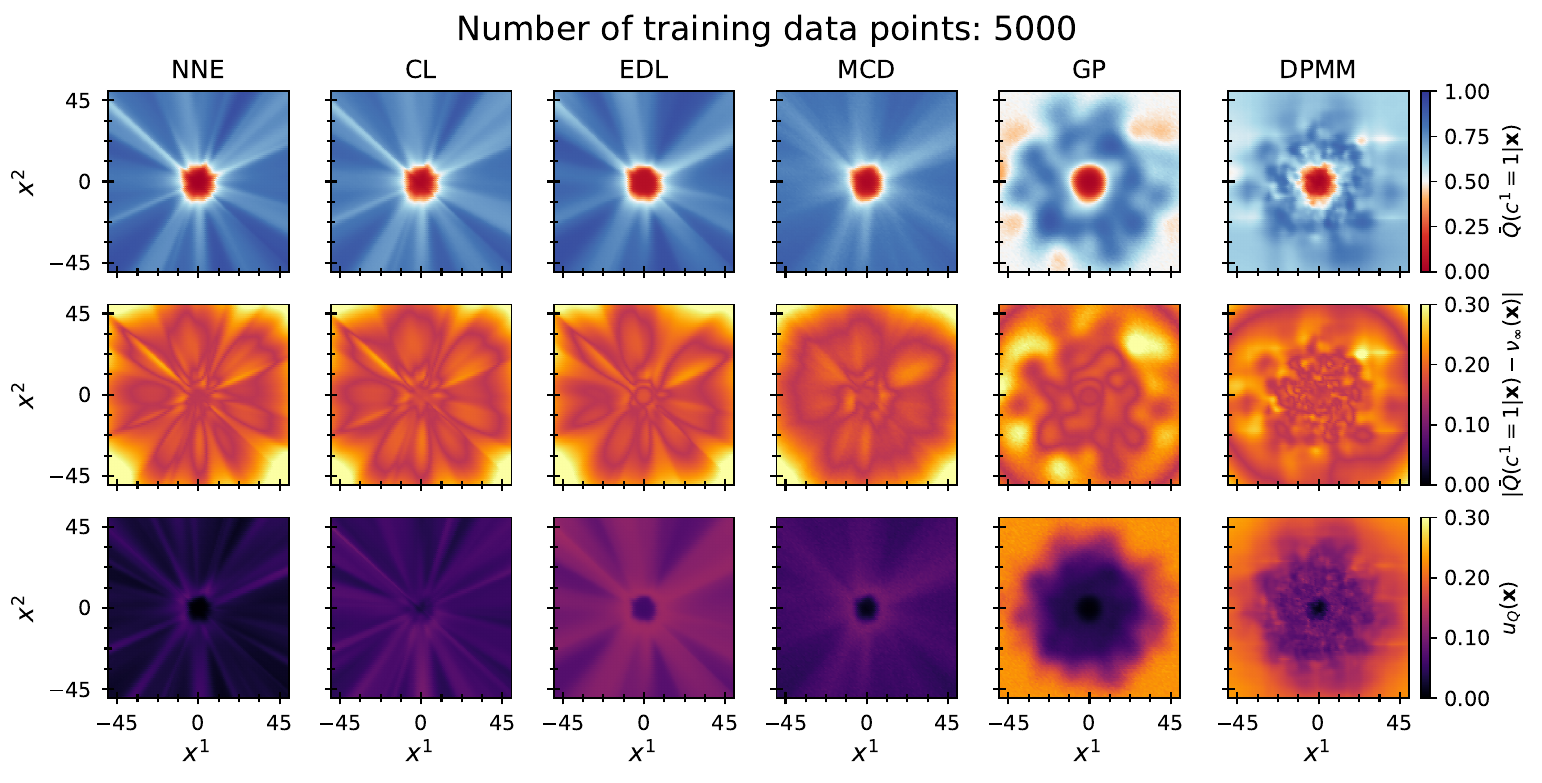}
    
    \caption{Estimated probabilities (top row), the absolute difference between the LRFD and the estimated probabilities (middle row) and uncertainties (bottom row) for $N_\text{train}=5000$ using the \texttt{ReLU} activation function (top) and the \texttt{tanh} activation function (bottom) for the deep learning algorithms in the study for dataset A.}
    \label{fig:gridplots}
\end{figure}

Figure \ref{fig:gridplots} shows the spatial distribution of probability estimates, difference between the estimated distributions and the LRFD and the estimated uncertainties and illustrates the impact of changing the activation function in the deep learning architectures. These visualizations nicely illustrate the impact of the inherent spatial structures of the different models. The  \texttt{ReLU} neural network models like NNE and EDL produce distributions with linear structure. MCD blurs the linear structure, creating a smoother distribution. The linear structure is even more prominent when using the \texttt{tanh} activation function. GP with RBF kernel produces relatively smooth distributions with local radial symmetry. The specific DPMM kernels used in this study are axes-parallel bivariate Gaussians, which give the distinct grid-like structure seen in the plots. These results highlight the importance of choosing appropriate model architecture (or transforming the data to fit the model), even when the modeling tools are in theory very flexible. 

\end{document}